\documentclass{article}

\usepackage{authblk}
\usepackage{PRIMEarxiv}
\usepackage{amsmath,amsbsy,amsfonts,amssymb,amsthm,bm}
\usepackage[utf8]{inputenc} 
\usepackage[T1]{fontenc}    
\usepackage{hyperref}       
\usepackage{url}            
\usepackage{booktabs}       
\usepackage{amsfonts}       
\usepackage{nicefrac}       
\usepackage{microtype}      
\usepackage{lipsum}
\usepackage{fancyhdr}       
\usepackage{graphicx}       
\usepackage{colortbl}
\usepackage{xcolor}
\usepackage{amsmath}
\usepackage{url}
\usepackage[title]{appendix}
\usepackage{booktabs}    
\usepackage{makecell}    
\usepackage{geometry}
\usepackage{subcaption}
\usepackage{tcolorbox} 
\usepackage{xcolor}
\usepackage{listings}
\tcbuselibrary{skins} 
\tcbuselibrary{breakable}
\tcbuselibrary{skins}

\newtcolorbox{dialoguebox}{
    colback=blue!5!white, 
    colframe=blue!50!black, 
    sharp corners, 
    boxrule=1pt, 
    fonttitle=\bfseries, 
    left=3mm, 
    right=3mm, 
    top=3mm, 
    bottom=3mm 
}
\graphicspath{{media/}}     
\makeatletter
\def\@fnsymbol#1{\ensuremath{\ifcase#1\or \or \dagger\or \ddagger\or
   \mathsection\or \mathparagraph\or \|\or **\or \dagger\dagger
   \or \ddagger\ddagger \else\@ctrerr\fi}}
\makeatother

\title{MUA-RL: Multi-turn User-interacting Agent Reinforcement Learning for agentic tool use}

\author{
Weikang Zhao\textsuperscript{1,2,*}\quad
Xili Wang\textsuperscript{1,3,*,$\dagger$}\quad
Chengdi Ma\textsuperscript{3} \quad
Lingbin Kong\textsuperscript{1} \quad \\
Zhaohua Yang\textsuperscript{1} \quad
Mingxiang Tuo\textsuperscript{1} \quad
Xiaowei Shi\textsuperscript{1} \quad
Yitao Zhai\textsuperscript{1,$\dagger$}\quad
Xunliang Cai\textsuperscript{1}\\
\\
\textsuperscript{1}Meituan Inc. \quad
\textsuperscript{2}Institute of Automation, Chinese Academy of Sciences \quad
\textsuperscript{3}Peking University \\
\\
\texttt{zhaoweikang2023@ia.ac.cn} \quad
\texttt{\{xiliwang,mcd2020\}@stu.pku.edu.cn} \\
\texttt{\{konglingbin,yangzhaohua02,tuomingxiang02,shixiaowei02,zhaiyitao,caixunliang\}@meituan.com}
\thanks{* Weikang Zhao and Xili Wang contributed equally to this work. \quad $\dagger$ Corresponding authors.}
}

\begin{document}
\maketitle

\begin{abstract}

With the recent rapid advancement of Agentic Intelligence, agentic tool use in LLMs has become increasingly important. During multi-turn interactions between agents and users, the dynamic, uncertain, and stochastic nature of user demands poses significant challenges to the agent's tool invocation capabilities. Agents are no longer expected to simply call tools to deliver a result; rather, they must iteratively refine their understanding of user needs through communication while simultaneously invoking tools to resolve user queries. Existing reinforcement learning (RL) approaches for tool use lack the integration of genuinely dynamic users during the RL training process. To bridge this gap, we introduce MUA-RL (Multi-turn User-interacting Agent Reinforcement Learning for agentic tool use), a novel reinforcement learning framework that, for the first time in the field of agentic tool use, integrates LLM-simulated users into the reinforcement learning loop. MUA-RL aims to enable autonomous learning of models to communicate with users efficiently and use various tools to solve practical problems in dynamic multi-turn interactions. Evaluations are done on several multi-turn tool-using benchmarks (see Figure~\ref{fig:charts}). Specifically, MUA-RL-32B achieves 67.3 on TAU2 Retail, 45.4 on TAU2 Airline, 28.3 on TAU2 Telecom, 28.4 on BFCL-V3 Multi Turn, and 82.5 on ACEBench Agent -- outperforming or matching the performance of larger open-source models such as DeepSeek-V3-0324 and Qwen3-235B-A22B in non-thinking settings.
\end{abstract}


\begin{figure}[htbp]
  \centering
  \includegraphics[width=1.0\textwidth]{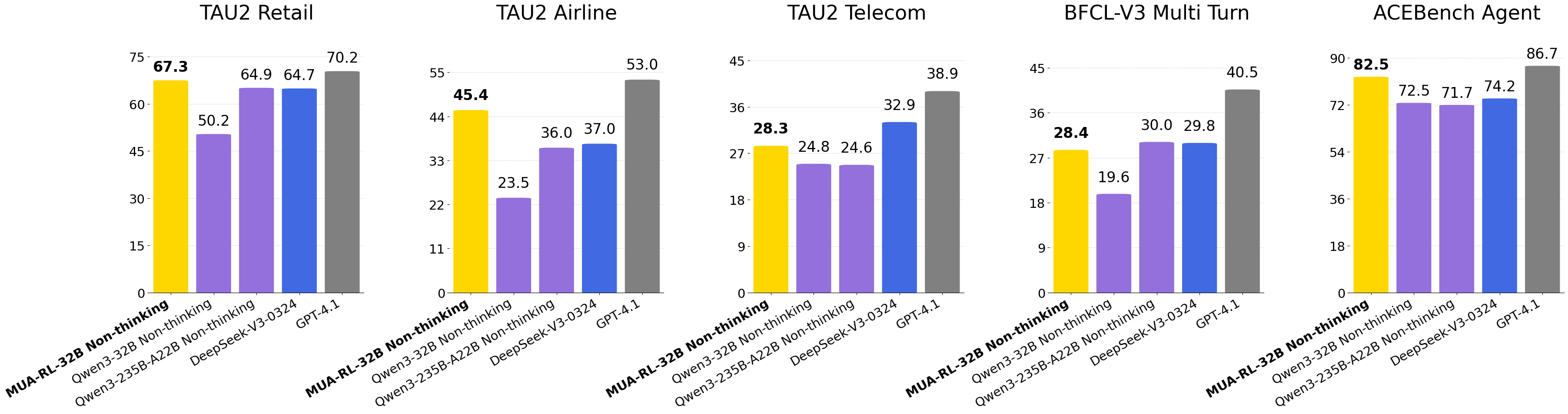}
  \caption{MUA-RL main results.}
  \label{fig:charts}
\end{figure}

\section{Introduction}

The paradigm of Large Language Models (LLMs) is shifting towards Agentic Intelligence -- models are now equipped with extensive external tools to interact dynamically with users and the world. This shift signifies a move away from static imitation learning, paving the way for models that engage in active learning through interactions, going beyond their pre-training and post-training data. Consequently, this new paradigm demands more on the model's ability to use tools effectively in multi-turn interactions and communicate clearly with users. While current LLMs mainly undergo supervised fine-tuning (SFT) on synthesized tool-using data to acquire capabilities to interact with both users and the world, reinforcement learning (RL) is believed to have better generalization than SFT~\cite{team2025kimi}.

Combined with RL, Large language models (LLMs) have shown remarkable progress, particularly in static, well-defined domains. For example, DeepSeek-Math~\cite{shao2024deepseekmath} achieves strong performance in mathematical problem solving. Recent works have begun to incorporate interactions with external environments into the reinforcement learning process of LLMs. Retool~\cite{feng2025retool} enhances RL by integrating code interpreter to enable LLMs to interact with a real code sandbox, SkyRL~\cite{cao2025skyrl} facilitates interactions between LLMs and actual Docker environments during RL training, and RAGEN~\cite{wang2025ragen} depends on symbolic or pre-specified environments. However, these systems generally operate within predetermined environments and rely on pre-scripted queries. As a result, current RL approaches may struggle on the dynamics and unpredictability of real-world interactions with users.

In real-world scenarios, users' needs are highly dynamic and unpredictable. Users often adjust their questions and expectations based on the model's responses, creating a feedback loop that requires continuous bidirectional adaptation. This interactive and co-evolutionary dynamic is currently overlooked in existing RL training frameworks and remains largely unexplored in practice. 
\textit{How can we design reinforcement learning frameworks for LLMs that capture dynamic, user-driven interactions, enable models to adapt in real time to evolve goals and feedback?}

To investigate the impact of dynamic user roles during the reinforcement learning rollouts, we propose MUA-RL, a novel reinforcement learning framework that, for the first time in the field of agentic tool use, integrates LLM-simulated users into the reinforcement learning loop. MUA-RL is designed to employ end-to-end reinforcement learning to enhance the agent’s ability to iteratively refine its understanding of user intent through ongoing dialogue, while actively invoking tools to fulfill user requests. Unlike approaches that incentivize specific tool-calling formats or success rates at intermediate steps, our framework provides reward solely based on ultimate task completion. This encourages more bold and diverse exploration during reinforcement learning processes, fostering the emergence of more robust and general behavioral patterns. Ultimately, the agent autonomously evolves its capability to handle multi-turn tool-using tasks.

Our main contributions are as follows:

\begin{itemize}
    \item We proposed MUA-RL, a novel multi-turn user-interacting reinforcement learning framework that incorporates LLM-simulated users into the reinforcement learning rollouts, and integrated a real, operational database environment for validating the results generated by tool invocation. The framework will be open-sourced on \url{https://github.com/zzwkk/MUA-RL}.
    \item To develop the model's basic capability to invoke tools, we constructed two agentic data synthesis pipelines for high-quality cold-start training: one in which tool responses are simulated by LLMs, and another where tool responses are provided by real MCP server. The high-quality cold-start data will be open-sourced on \url{https://huggingface.co/datasets/zzwkk/MUA-RL-Dataset}.
    \item We conducted detailed analysis of the model’s training dynamics throughout the multi-turn user-interacting reinforcement learning process, as well as a comparative examination of its behavior before and after the application of MUA-RL. All MUA-RL models will be open-sourced on \url{https://huggingface.co/zzwkk/MUA-RL-8B}, \url{https://huggingface.co/zzwkk/MUA-RL-14B}, \url{https://huggingface.co/zzwkk/MUA-RL-32B}.
\end{itemize}



\section{Related Work}
\label{sec:related work}

\paragraph{Reinforcement learning.} 

Recent progress in reinforcement learning (RL) for LLMs has moved beyond classical algorithmic frameworks toward more scalable and efficient optimization paradigms. Early work primarily leveraged Proximal Policy Optimization (PPO)~\cite{schulman2017proximal} and actor–critic approaches~\cite{haarnoja2018soft} to align models with human-preferred behaviors. Subsequent developments have introduced policy variants such as RLOO~\cite{Kool_Hoof_Welling_2019}, GRPO~\cite{shao2024deepseekmath}, and DAPO~\cite{yu2025dapo}, which aim to improve stability and sample efficiency under large-scale training. In parallel, STaR, through its iterative self-derivation process~\cite{zelikman2022star}, and MCTS, via its lookahead-based exploration~\cite{hao2023reasoning}, have both demonstrated the ability to significantly reduce reliance on external supervision while substantially improving sample efficiency. More recently, RL has been integrated into broader LLM fine-tuning frameworks, enabling models to adapt flexibly to diverse downstream tasks such as search~\cite{jin2025search}, coding~\cite{liu2025code}, and multimodal tasks~\cite{shen2025vlm}. These advances underscore the versatility of RL as a general optimization framework for enhancing LLM capabilities, providing a foundation for extending RL techniques into new domains of model alignment and task specialization.

\paragraph{Agentic tool use.}

The development of tool use in LLMs has progressed through a series of methodological stages. Initially, prompting-based approaches relied on carefully designed instructions to elicit tool invocation without additional training~\cite{chen2022program, lei2023instructerc}. While lightweight and flexible, these methods often exhibited unstable tool behavior and struggled with complex tool interactions. Subsequently, supervised fine-tuning (SFT) methods emerged, where models were trained on curated or distilled trajectories to improve their accuracy and consistency in tool use~\cite{lin2024hammer, zhang2025xlam, acikgoz2025can}. These approaches achieved more reliable performance but were constrained by the coverage and quality of training data. More recently, reinforcement learning (RL) techniques have been employed to optimize tool-using policies through outcome-driven feedback~\cite{song2025r1, sun2025zerosearch, li2025torl, singh2025agentic}, allowing models to explore and refine tool invocation strategies in interactive environments. Despite these advances, most existing work centers on text-based or multi-step tasks (Section~\ref{sec:multi-turn user}), and challenges remain in scaling to dynamic multi-turn user-interacting tool use.






\section{Methodology}
\label{sec:meth}

\subsection{Task Formulation}

\label{sec:task_defin}

Multi-turn tool-using tasks involve dynamic interactions between the user and the agent and extensive exchanges between the database and the agent. 
At each interaction turn, the agent may: (a) invoke one or more tools sequentially to interact with the database for information retrieval or operations, 
\textit{or} (b) communicate textually with the user to acquire information and discern their intent.

\begin{figure}[htbp]
  \centering
  \includegraphics[width=0.9\textwidth]{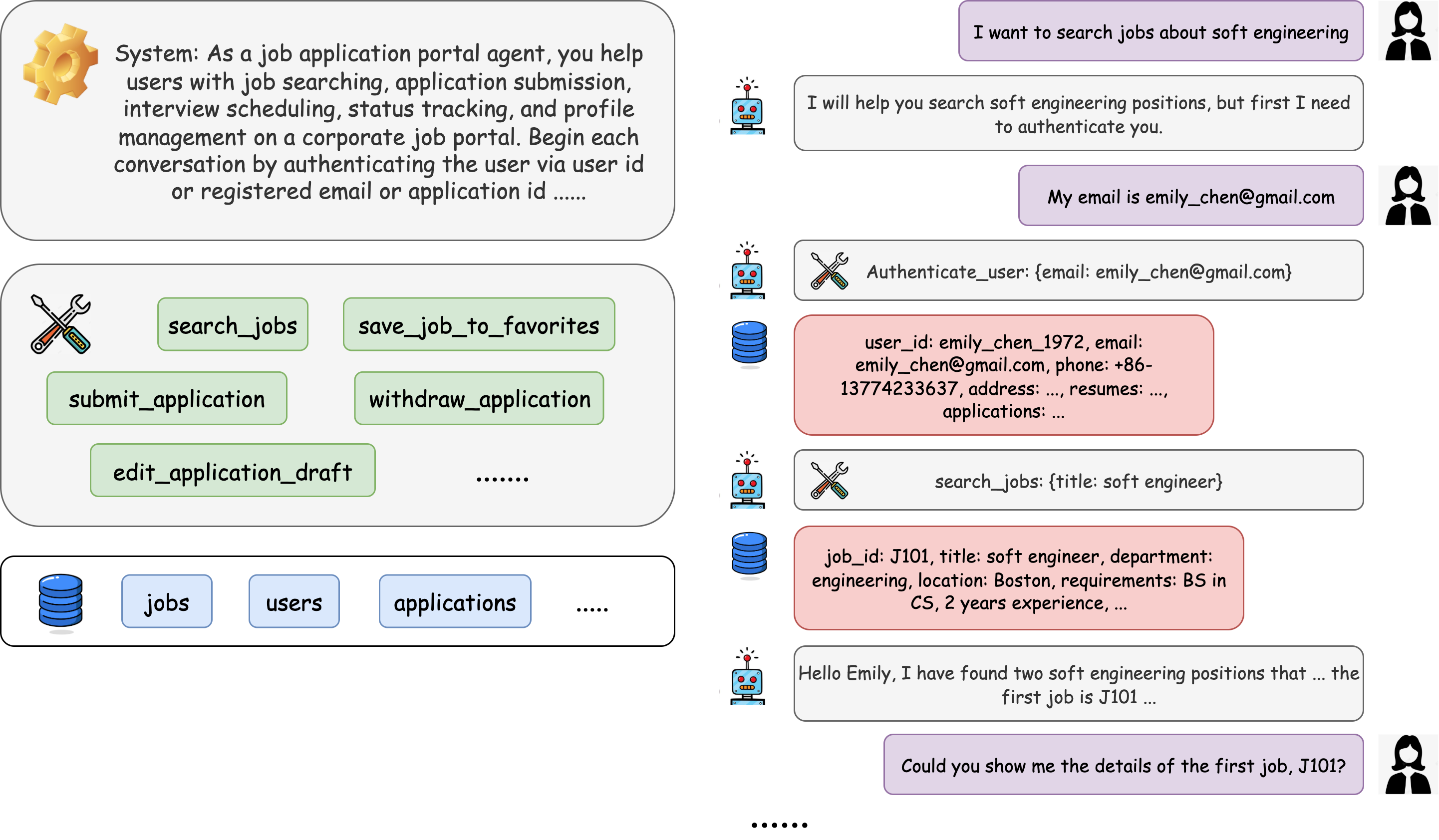}
  \caption{A trajectory example in multi-turn tool-using tasks. Here, the agent invokes tools to interact with the database, or communicates textually with the user.}
  \label{fig:multiturn_illus}
\end{figure}

Formally, define a tuple $(\mathcal{T},\mathcal{M},\mathcal{O})$ where:
$\mathcal{T}$ is the tool set space,
$\mathcal{M}$ is the message space,
$\mathcal{O} = \mathcal{O}_{\text{db}} \cup \mathcal{O}_{\text{user}}$ is the observation space,
with $\mathcal{O}_{\text{db}}$ and $\mathcal{O}_{\text{user}}$ denoting the database and user observation subspaces respectively.
Given a user query $o_{1,\text{user}}\in \mathcal{O}_{\text{user}}$, a typical multi-turn trajectory is expressed as:
\begin{equation}
\underbrace{(o_{1,\text{user}}\to t_1\to o_{1,\text{db}}\to...\to m_1)}_{\text{turn 1}},...,\underbrace{(o_{k,\text{user}}\to t_j\to o_{j,\text{db}}\to ...\to t_{j+j_k}\to o_{j+j_k,\text{db}}\to m_k)}_{\text{turn $k$}},...
\end{equation}
where $t_i\in\mathcal{T}$ denotes invoking tool, $o_{i,\text{db}}$ denotes observation from database after invoking $t_i$, $o_{i,\text{user}}$ is the observation from the user, and $m_i\in\mathcal{M}$ represents the agent's message to the user. 
At each turn, the agent autonomously decides \textit{whether} to invoke tools and \textit{how many} tools to invoke, in order to make progress towards solving the user query. 
A multi-turn tool-using trajectory example is illustrated in Figure~\ref{fig:multiturn_illus}.

\subsection{Agentic Data Synthesis Pipeline for High-quality Cold-start}
\label{sec:csta}

The multi-turn tool-using tasks introduced in Section~\ref{sec:task_defin} present significant challenges for language models. The agent must autonomously invoke unfamiliar tools, and iteratively act through text-based communication, tool invocation, and error correction. 
To address these challenges, we employ a lightweight supervised fine-tuning phase (cold-start) to establish the basic capabilities of models in handling multi-turn tool-using tasks prior to deploying them within a reinforcement learning framework for self-iteration.

Although the real world provides rich and authentic interaction scenarios, conducting data collection in such settings is difficult due to cost limitations, system complexity, privacy concerns, and accessibility barriers. Recent research efforts~\cite{mitra2024agentinstruct,guo2024stabletoolbench,team2025kimi,sun2025zerosearch} have focused on synthetic approaches for generating tool-using data, where large language models (LLMs) are typically employed to simulate tool execution results. Notably, the emergence of Model Context Protocol (MCP)~\cite{hou2025model} now offers an alternative approach, enabling direct utilization of accessible MCP servers for real tool execution.

\paragraph{Tool execution simulated by Large Language Models (LLMs).}

\begin{figure}[htbp]
  \centering
  \includegraphics[width=1.0\textwidth]{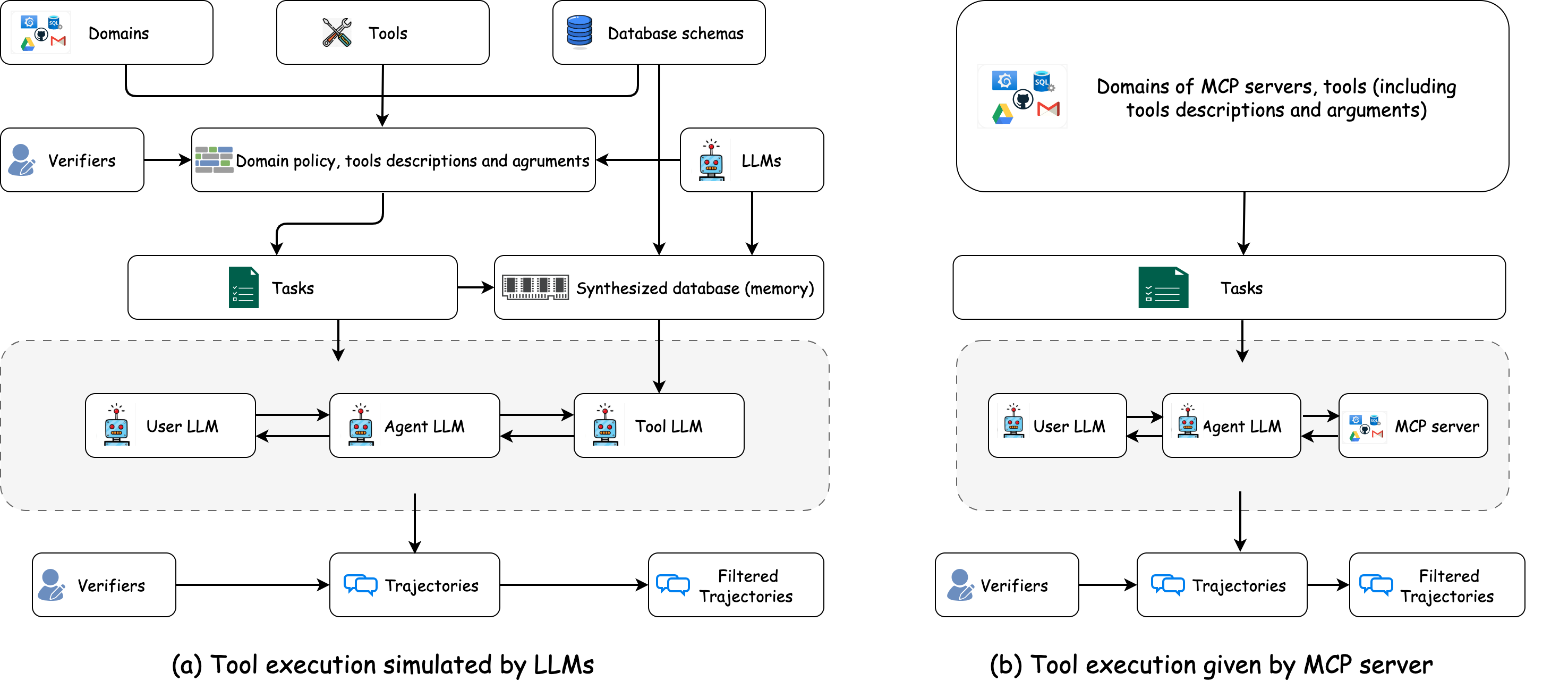}
  \caption{Agentic data synthesis pipeline. (a) Tool execution results simulated by LLMs. (b) Tool execution results given by real MCP server.}
  \label{fig:llm-mcp}
\end{figure}

In this scenario, we first come out a domain and its simplest possible database schemas, along with corresponding tools, inspired by (and simplified from) its real-world counterpart. Next, descriptions and arguments of the tools, and domain policy are generated by LLMs, followed by human curation and iterative refinement. Then, we generate trajectories through the collaboration of three LLMs: one serving as the \textit{agent}, another as the \textit{user}, and--most critically--one as the \textit{tool}. While the roles of the agent and user are relatively straightforward, the tool LLM requires careful design to ensure the reliability of the whole generation process. Specifically, when constructing a domain-specific query, we employ LLMs to generate a small synthetic database (serving as memory) that conforms to the predefined database schemas. This memory is then provided to the tool LLM. During trajectory generation, when the agent LLM invokes a certain tool, it passes the tool name and corresponding arguments to the tool LLM. The tool LLM subsequently processes this input and generates an appropriate tool response based on the provided memory, as shown in Figure~\ref{fig:llm-mcp}(a).

\paragraph{Tool execution given by real Model Context Protocol (MCP) server.}

In the MCP scenario, the process is significantly simplified. All tools are presented upfront, eliminating the need for manual design, and the MCP server automatically handles all tool execution. What remains is generating domain-specific user queries corresponding to the MCP server and producing trajectories through the interactions between the agent LLM, user LLM, and the MCP server, as illustrated in Figure~\ref{fig:llm-mcp}(b).

It is noteworthy that all cold-start datasets undergo dual-verification, which combines human expert annotation with DeepSeek-R1 evaluation, to filter out invalid trajectories and ensure data quality and validity. Trajectory examples are provided in Appendix~\ref{appen:csd}.

\subsection{MUA-RL: Multi-turn User-interacting Agent Reinforcement Learning for agentic tool use}
\label{sec:MUA-rl}

\subsubsection{Group Relative Policy Optimization (GRPO)} 
\label{sec:MUA-rl-grpo}

Following recent advances in reinforcement learning (RL) for LLMs, we adopt Group Relative Policy Optimization (GRPO)~\cite{shao2024deepseekmath} -- a sample-efficient algorithm that optimizes policies directly via groupwise outcome rewards. GRPO’s elimination of value function approximation reduces training complexity while maintaining stability, as evidenced by its success in DeepSeek-R1~\cite{guo2025deepseek}. The reduced training cost and simplified optimization steps make GRPO a suited RL algorithm for our multi-turn user-interacting training.

Specifically, given an existing old policy $\pi_{\text{old}}$, a reference policy $\pi_{\text{ref}}$, a group of responses $\{y_1,y_2,...,y_G\}$ is sampled from the old policy $\pi_{\text{old}}$ for each query $q$. The policy model $\pi_{\bm{\theta}}$ is then optimized by maximizing the following objective,

\begin{equation}
\label{grpo}
\begin{aligned}
\mathcal{J}_{\text{GRPO}}(\bm{\theta}) := \mathbb{E}_{q\sim \mathcal{D},\{y_i\}_{i=1}^G\sim\pi_{\text{old}}(\cdot|q)} \frac{1}{G}\sum_{i=1}^G\Bigg(&\min\left(\frac{\pi_{\bm{\theta}}(y_i|q)}{\pi_{\text{old}}(y_i|q)}A_i,\text{clip}\left(\frac{\pi_{\bm{\theta}}(y_i|q)}{\pi_{\text{old}}(y_i|q)},1-\epsilon,1+\epsilon\right)A_i\right)\\
&-\beta\mathbb{D}_{\text{KL}}(\pi_{\bm{\theta}}\|\pi_{\text{ref}})\Bigg),
\end{aligned}
\end{equation}
where $\epsilon$ and $\beta$ are hyper-parameters, and $A_i$ is the advantage computed using a group of rewards $\{r_1,r_2,...,r_G\}$ corresponding to responses $\{y_1,y_2,...,y_G\}$:
\begin{equation}
A_i := \frac{r_i-\text{mean}(\{r_{i}\}_{i=1}^G)}{\text{std}(\{r_{i}\}_{i=1}^G)}.
\end{equation}

\subsubsection{Multi-turn User-interacting Rollout with Real-time Tool Execution} 
\label{sec:multi-turn user}

\begin{figure}[htbp]
  \centering
  \includegraphics[width=1.00\textwidth]{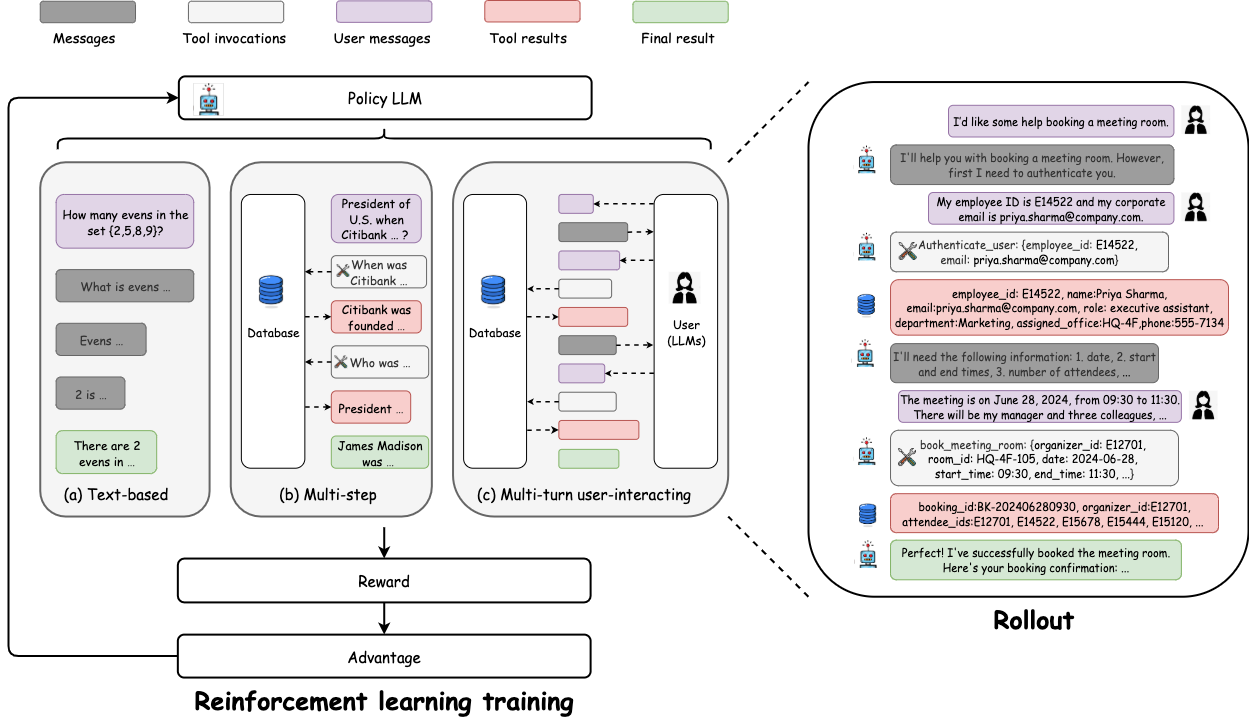}
  \caption{Three different kinds of the rollout processes. (a) The text-based rollout. (b) The multi-step rollout with tool execution. (c) The multi-turn user-interacting rollout with tool execution.}
  \label{fig:MUA_illus}
\end{figure}

In conventional rollout processes, the policy LLMs typically perform only text-based generation, and then output a final answer to obtain a reward (e.g., in most mathematical reasoning tasks~\cite{shao2024deepseekmath, seed2025seed1}). During inference, they do not use tools to interact with external systems--even commonly used ones like code interpreter (CI)--as illustrated in Figure~\ref{fig:MUA_illus}(a).

Recent work explores integrating text generation with executable tools in rollout processes.
As demonstrated in Figure~\ref{fig:MUA_illus}(b), these approaches dynamically interleave natural language generation by the policy LLM with real-time tool-execution results on the fly. Here, the term "database" can carry different meanings depending on the scenarios. 
With a CI, the "database" may denote a real code-execution sandbox environment~\cite{li2025torl, feng2025retool}, whereas in deep research settings it can refer to the internet as an information-retrieval source~\cite{jin2025search, sun2025zerosearch, song2025r1, chen2025learning, li2025webthinker}.
This interactive rollout paradigm (also called multi-step rollout), which engages with real-world databases, enables the model to develop practical tool invocation capabilities for solving domain-specific problems, moving beyond pure textual generation.

Advancing further, recent works have explored using LLMs as simulators of human characters~\cite{kim2022plm, park2023generative}, which opens new possibilities for integrating user roles in large-scale reinforcement learning rollout processes. Building upon this, in our reinforcement learning framework, we introduce automated users simulated by LLMs to interact with the agent during rollouts.
Compared to the text-based rollout and multi-step rollout, the agent must not only communicate with the user via text to gather user information and discern user's intentions, but also utilize provided tools to interact with the database -- retrieving information or performing operations to fulfill the user's requests. The complete rollout integrates multiple components: text-based generation, tool invocation, user messages, and tool-execution results, which significantly enhances the dynamics, stochasticity, and uncertainty of the rollout processes. Through this highly dynamic framework, we aim to train agents to autonomously develop more sophisticated behavioral patterns, requiring both appropriate tool usage and effective communication with users, as depicted in Figure~\ref{fig:MUA_illus}(c).

\subsubsection{Reward} 

In existing reinforcement learning works on agentic tool use, the reward design is often quite complex. Since tool invocation is a highly structured behavior for language models, reward design typically incorporates format rewards~\cite{singh2025agentic, qian2025toolrl, feng2025retool}, 
tool-name or parameter-name matching rewards~\cite{qian2025toolrl, zhang2025nemotron},
and tool execution rewards that measure fraction of successful tool calls~\cite{singh2025agentic, zeng2025reinforcing}. However, complex reward design may not enable models to learn effective behavioral patterns in dynamic multi-turn interactions. Instead, the model may becomes discouraged from trial and error. In our work, we simplify the reward design, the reward $r=1$ only when the agent successfully fulfills the task in accordance with the system prompt, and $r=0$ otherwise. This reward design offers two advantages: 

\begin{itemize}
    \item \textbf{Robustness to dialogue variation:} The evaluation is invariant to the specific conversational trajectory or tool invocation sequence, allowing for diverse agent behaviors so long as the correct outcome is achieved.
    \item \textbf{Mitigation of reward hacking:} Agents cannot directly exploit the output format or tool invocation syntax but only be rewarded by complete task resolution.
\end{itemize}

\section{Experiments}
\label{sec:experiments}

In this section, we demonstrate the superiority of MUA-RL in terms of performance, robustness, and generalization across multiple multi-turn tool-using benchmarks, and conduct in-depth analyses to verify the effectiveness of our multi-turn user-interacting reinforcement learning framework.

\subsection{Experiment Setup}

\paragraph{Models.} We use Qwen3-8B Non-thinking, Qwen3-14B Non-thinking, and Qwen3-32B Non-thinking~\cite{yang2025qwen3} as the primary backbone models in our experiments. These models are selected to cover a range of parameter scales while maintaining architectural consistency, enabling us to analyze scaling effects.

\paragraph{Cold-start.} We synthesized approximately two thousand trajectories for cold-start training using the agentic data synthesis pipeline described in Section~\ref{sec:csta}. These trajectories span 9 scenarios, including 5 synthetic scenarios and 4 real-world MCP server scenarios. For training hyperparameters, the models are trained with a batch size of 128 for 2 epochs using the AdamW optimizer~\cite{loshchilov2017decoupled}, an initial learning rate of 5e-6, and a cosine decay learning rate schedule.

\paragraph{Reinforcement learning training.} We implemented multi-turn user-interacting reinforcement learning framework based on VolcEngine Reinforcement Learning (VeRL)~\cite{sheng2024hybridflow} and integrated a real, operational database environment for validating the results generated by tool invocation. The framework will be released on GitHub\footnote{\url{https://github.com/zzwkk/MUA-RL}}. For training data, 115 retail and 50 airline datasets from \textbf{TAU1-Bench}~\cite{yao2024tau} were used. TAU1-Bench is a benchmark simulating realistic multi-turn user-agent interactions, requiring domain-specific tool use and adherence to policy constraints. During RL training, we simplified the reward computation in TAU1-Bench datasets. Originally, models received reward $r=0$ if they either failed to complete the task or failed to mention specific required information in dialogue (e.g., telling the user how many clothing items are in stock). We removed the dialogue content requirements, so models now only need to successfully complete the task to receive reward $r=1$. This adjustment is necessitated by our empirical observation that dialogue content requirements impede the model's ability to learn correct tool invocation patterns. For RL training algorithm, Group Relative Policy Optimization (GRPO) algorithm introduced in Section~\ref{sec:MUA-rl-grpo} is adopted, where Kullback-Leibler (KL) loss coefficient $\beta=0.001$. The training configuration employs 25 epochs with a batch size of 32 and a rollout number of 8. GPT-4o-2024-11-20~\cite{hurst2024gpt} is used as the LLM user simulator. The sequence length is limited to 32768 tokens. Furthermore, we establish an upper bound of 30 interaction turns per task to ensure computational efficiency and prevent excessive interaction turns. The temperature of the agent during rollout is set to $1.0$.

\paragraph{Loss mask strategy.}  To enable the model to learn effective tool invocation strategies and efficient communication patterns, we implemented a loss mask strategy that masks tokens from tool execution results and user messages during loss computation.

\paragraph{Benchmarks.} We evaluate our approach on four representative multi-turn tool-using benchmarks. In addition to the previously introduced TAU1-Bench, three other multi-turn tool-using benchmarks -- \textbf{TAU2-Bench}~\cite{barres2025tau}, \textbf{Berkeley Function-Calling Leaderboard (BFCL)-V3 Multi Turn}~\cite{patilberkeley}, and \textbf{ACEBench Agent}~\cite{chen2025acebench}, are evaluated to demonstrate the generalization of our method. Specifically, 
while maintaining the airline and retail domains, TAU2 differs from TAU1 through modified tool sets (with removals and additions), refined policies and reward mechanisms, stricter evaluation criteria, and introduces a new and more complex \textit{dual-control} domain, TAU2 Telecom, where both the agent and user can invoke tools. BFCL-V3 Multi Turn consists of a foundational Multi Turn Base test set, as well as three distinct types of augmented multi-turn test sets: Miss Param, Miss Func, and Long Context. ACEBench Agent encompasses four real fundamental scenarios, categorized into multi-step and multi-turn tasks. See Appendix~\ref{app:c} for further details.

\paragraph{Evaluation.} For TAU1-Bench and TAU2-Bench, we use GPT-4.1 as the user simulator. For each test set, we conduct four repeated tests and take the average to improve confidence. Besides, for TAU2 Telecom, we introduce the \textbf{Task Completion Rate} (\textit{TCR}) as a fine-grained evaluation metric for assessing the performance of agents. Unlike binary success/failure metrics, \textit{TCR} provides a continuous measurement of task completion by evaluating the proportion of satisfied verification criteria relative to the total required criteria for a given task. Formally, for a task $q$ with $n$ verification criteria, the \textit{TCR} is defined as:

$$
\textit{TCR}(q) := \frac{|\{c_i \in C_q : \text{satisfied}(c_i)\}|}{|C_q|},
$$

where $C_q = \{c_1, c_2, ..., c_n\}$ represents the set of verification criteria for task $q$, and $\text{satisfied}(c_i)$ indicates whether criterion $c_i$ is successfully met by the agent. The \textit{TCR} ranges from $0.0\%$ (complete failure) to $100.0\%$ (complete success), enabling more nuanced performance evaluation across different task complexities and agent capabilities. For BFCL-V3 Multi Turn, models are evaluated using Executable Function Accuracy, which evaluates whether the generated functions execute correctly and yield the expected outputs. For ACEBench Agent, the user simulator is the same as that in TAU-Bench. In addition, considering that our model and mainstream models typically invoke tools via Function Calling (FC) rather than prompts, all of our test sets are evaluated using the FC mode. To obtain reliable results, we compare our MUA-RL series against leading large language models such as GPT-4o~\cite{hurst2024gpt} and DeepSeek-V3~\cite{liu2024deepseek}. All evaluations are conducted under a deterministic inference setting with temperature fixed to $0.0$ and non-thinking mode enabled, ensuring reproducibility and eliminating randomness introduced by stochastic decoding.

\subsection{Performance on Various Benchmarks}


\subsubsection{Detail Performance on TAU1-Bench \& TAU2-Bench}

Table~\ref{tab:taubench} presents a comprehensive comparison of MUA-RL series with a range of baselines across different model scales. In TAU1-Bench and TAU2-Bench, the MUA-RL series demonstrate clear improvements over their base and cold-start counterparts. Notably, despite its smaller size, the MUA-RL-32B demonstrates highly competitive performance, not only matching the capabilities of GPT-4.1 but also surpassing much larger models such as Qwen3-235B-A22B, DeepSeek-V3-0324, and GPT-4o-2024-11-20 in TAU Retail and TAU Airline. Moreover, in the challenging TAU2 Telecom with dual-control dynamics, MUA-RL-14B achieves an accuracy of 33.4\%, surpassing the performance of GPT-4o-2024-11-20 and DeepSeek-V3-0324, highlighting that our approach is both robust and highly adaptable to complex, real-world scenarios.

While cold-start models show improvements on TAU Retail and TAU Airline, they exhibit degraded performance on TAU Telecom. However, MUA-RL models exhibit strong generalization capabilities, consistently achieving competitive or leading performance across all domains, especially on the more intricate telecom domain. The observed performance degradation of cold-start models on the telecom domain can be attributed to the introduction of domain-specific patterns and biases through the cold-start training data. While these patterns can provide advantages in areas similar to the training distribution, they struggle with generalization, particularly when encountering domains that differ greatly from the training data. In contrast, MUA-RL enables the models to effectively counteract biases introduced during the SFT process, guiding them towards more robust and generalizable behavioral patterns.

Furthermore, the \textit{TCR} metric in TAU2 Telecom reveals the effectiveness of our approach in partial task completion scenarios. MUA-RL-14B achieves a \textit{TCR} of $54.3\%$, higher than the cold-start model's $32.9\%$, indicating that even when the model fails to complete the full task, it demonstrates better partial completion capabilities.

\begin{table}[htbp]
\centering
\begin{tabular}{l cc cccc}
\toprule 
& \multicolumn{2}{c}{\textbf{TAU1}} 
& \multicolumn{4}{c}{\textbf{TAU2}} \\
\cmidrule(lr){2-3} \cmidrule(lr){4-7}
\textbf{Model}
& \textit{Retail} & \textit{Airline} 
& \textit{Retail} & \textit{Airline} & \textit{Telecom} & \textit{Telecom TCR} \\
\midrule
\multicolumn{7}{c}{Existing Baselines} \\
\midrule
GPT-4o-2024-11-20 & 63.0 & 45.5 & 67.3 & 46.9 & 24.1 & -- \\
GPT-4.1 & 66.5 & 42.5 & 70.2 & 53.0 & 38.9 & -- \\
DeepSeek-V3-0324 & 70.4 & 42.4 & 64.7 & 37.0 & 32.9 & -- \\
Qwen3-235B-A22B Non-thinking & 65.2 & 32.0 & 64.9 & 36.0 & 24.6 & -- \\
Qwen3-30B-A3B Non-thinking & 38.3 & 18.0 & 31.6 & 18.0 & 18.4 & -- \\
Qwen3-4B Non-thinking & 24.3 & 16.0 & 28.1 & 12.0 & 17.5 & -- \\
\midrule
\multicolumn{7}{c}{MUA-RL} \\ 
\midrule
Qwen3-8B Non-thinking & 40.0 & 11.0 & 41.0 & 12.5 & 19.1 & $22.9\%$ \\
Qwen3-8B Cold-start & 36.7 & 12.0 & 31.4 & 16.0 & 9.0 & $12.1\%$ \\
\rowcolor{lightgray}
MUA-RL-8B & 56.5 & 29.5 & 49.8 & 19.0 & 21.8 & $25.2\%$ \\
Qwen3-14B Non-thinking & 46.9 & 13.0 & 43.1 & 14.8 & 29.9 & $46.6\%$ \\
Qwen3-14B Cold-start & 50.8 & 23.0 & 53.7 & 24.0 & 23.5 & $32.9\%$ \\
\rowcolor{lightgray}
MUA-RL-14B  & 65.9 & 42.0 & 66.0 & 38.0 & 33.4 & $54.3\%$ \\
Qwen3-32B Non-thinking & 47.6 & 18.5 & 50.2 & 23.5 & 24.8 & $23.7\%$ \\
Qwen3-32B Cold-start & 58.9 & 36.0 & 58.2 & 31.1 & 19.3 & $21.6\%$ \\
\rowcolor{lightgray}
MUA-RL-32B  & 72.6 & 46.5 & 67.3 & 45.4 & 28.3 & $45.1\%$ \\
\bottomrule
\end{tabular}
\vspace{4pt}
\caption{Performance comparison of different models on TAU-Bench. Each model runs in non-thinking mode.}
\label{tab:taubench}
\end{table}


\subsubsection{Detail Performance on BFCL-V3 Multi Turn \& ACEBench Agent}

Table~\ref{tab:bfcl} showcases the performance of various models on BFCL-V3 Multi Turn and ACEBench Agent. In BFCL-V3 Multi Turn, the MUA-RL series exhibit steady performance gains across all model scales. Notably, MUA-RL-32B achieves an overall accuracy of 28.4\%, outperforming its base and cold-start models and approaching the performance of DeepSeek-V3-0324. Similarly, on ACEBench Agent, the MUA-RL models demonstrate consistently superior performance over their base and cold-start models. It's worth noting that MUA-RL-32B achieves a score of 82.5, which is the highest among all tested models except for GPT-4.1 (86.7). The above results from both BFCL-V3 Multi Turn and ACEBench Agent confirm the remarkable generalization ability of MUA-RL across diverse multi-turn tool-using tasks.

\begin{table}[htbp]
\centering
\begin{tabular}{l ccccc ccc}
\toprule
    & \multicolumn{5}{c}{\textbf{BFCL-V3 Multi Turn}} 
    & \multicolumn{3}{c}{\textbf{ACEBench Agent}} \\
\cmidrule(lr){2-6} \cmidrule(lr){7-9}
\textbf{Model}
    & \textit{Base}
    & \begin{tabular}{@{}c@{}}\textit{Miss}\\\textit{Func}\end{tabular}
    & \begin{tabular}{@{}c@{}}\textit{Miss}\\\textit{Param}\end{tabular}
    & \begin{tabular}{@{}c@{}}\textit{Long}\\\textit{Context}\end{tabular}
    & \begin{tabular}{@{}c@{}}\textit{Overall}\\\textit{Acc}\end{tabular}
    & \begin{tabular}{@{}c@{}}\textit{Multi}\\\textit{Turn}\end{tabular}
    & \begin{tabular}{@{}c@{}}\textit{Multi}\\\textit{Step}\end{tabular}
    & \begin{tabular}{@{}c@{}}\textit{Overall}\\\textit{Acc}\end{tabular} \\
\midrule
\multicolumn{9}{c}{Existing Baselines} \\
\midrule
GPT-4.1
    & 48.0
    & 34.0
    & 35.0
    & 45.5
    & 40.5
    & 83.3
    & 90.0
    & 86.7 \\
DeepSeek-V3-0324
    & 41.0
    & 21.0
    & 23.0
    & 34.5
    & 29.8
    & 73.3
    & 75.0
    & 74.2 \\
Qwen3-235B-A22B Non-thinking
    & 42.5
    & 23.5
    & 28.5
    & 25.5
    & 30.0
    & 63.3
    & 80.0
    & 71.7 \\
Qwen3-30B-A3B Non-thinking
    & 14.0
    & 1.5
    & 7.5
    & 8.5
    & 7.9
    & 36.7
    & 30.0
    & 33.4 \\
\midrule
\multicolumn{9}{c}{MUA-RL} \\
\midrule
Qwen3-8B Non-thinking
    & 20.0
    & 4.0
    & 13.0
    & 10.0
    & 11.8
    & 33.3
    & 45.0
    & 39.2 \\
Qwen3-8B Cold-start
    & 24.0
    & 11.0
    & 16.5
    & 10.0
    & 15.4
    & 36.7
    & 55.0
    & 45.9 \\
\rowcolor{lightgray}
MUA-RL-8B
    & 21.0
    & 11.5
    & 15.0
    & 11.0
    & 14.6
    & 46.7
    & 60.0
    & 53.3 \\
Qwen3-14B Non-thinking
    & 30.0
    & 8.0
    & 16.0
    & 16.5
    & 17.6
    & 40.0
    & 80.0
    & 60.0 \\
Qwen3-14B Cold-start
    & 35.0
    & 13.5
    & 21.5
    & 19.5
    & 22.4
    & 50.0
    & 90.0
    & 70.0 \\
\rowcolor{lightgray}
MUA-RL-14B
    & 40.5
    & 14.0
    & 25.0
    & 21.5
    & 25.3
    & 56.7
    & 100.0
    & 78.3 \\
Qwen3-32B Non-thinking
    & 29.5
    & 11.0
    & 20.0
    & 18.0
    & 19.6
    & 60.0
    & 85.0
    & 72.5 \\
Qwen3-32B Cold-start
    & 35.0
    & 21.0
    & 28.5
    & 19.5
    & 26.0
    & 53.3
    & 100.0
    & 76.6 \\
\rowcolor{lightgray}
MUA-RL-32B
    & 42.0
    & 20.0
    & 30.0
    & 21.5
    & 28.4
    & 70.0
    & 95.0
    & 82.5 \\
\bottomrule
\end{tabular}
\vspace{4pt}
\caption{Performance comparison of different models on BFCL-V3 Multi Turn and ACEBench Agent. Each model runs in non-thinking mode.}
\label{tab:bfcl}
\end{table}

\subsection{Deep Analysis}

\subsubsection{Training Dynamics} 

\begin{figure}[htbp]
  \centering
  \includegraphics[width=1.0\textwidth]{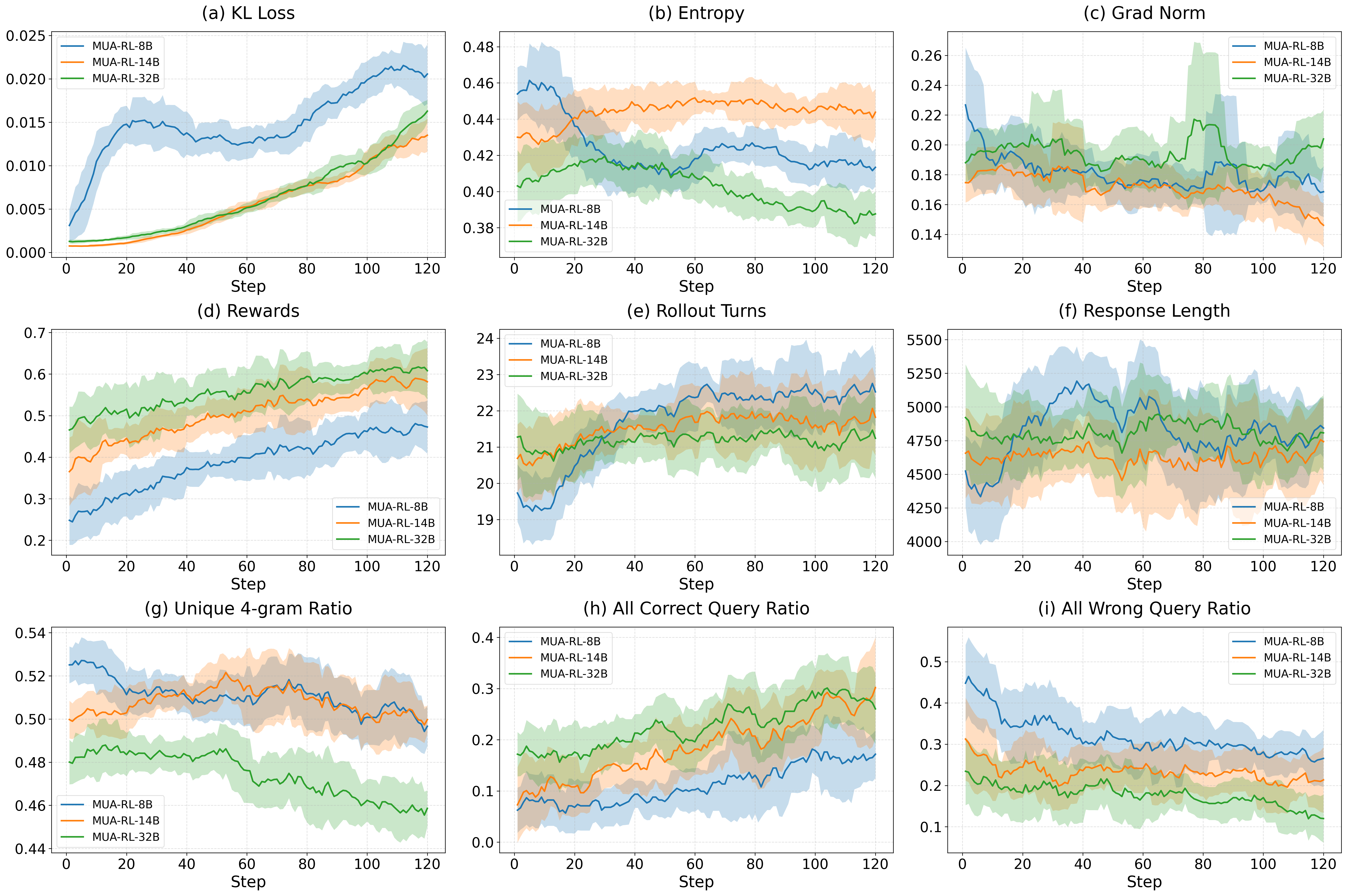}
  \caption{Learning curves of MUA-RL series during RL training.}
  \label{fig:batch_curve}
\end{figure}

This section presents visualizations of training dynamics of MUA-RL series and detailed analysis to gain further insights into MUA-RL. The corresponding learning curves are shown in Figure~\ref{fig:batch_curve}.

\paragraph{KL loss \& entropy \& grad norm.} The KL loss (Figure~\ref{fig:batch_curve}(a)) increases as the models gradually deviate from the cold-start models during RL training. Notably, MUA-RL-8B exhibits substantially larger fluctuations compared to MUA-RL-14B and MUA-RL-32B. We attribute this instability during the trade-off between exploration and regularization to the limited capacity of model with fewer parameters. In contrast, owning stronger representational power, larger models effectively average out noisy updates, leading to smoother curves of deviation. Moreover, the entropy curve (Figure~\ref{fig:batch_curve}(b)) indicates that MUA-RL-8B experienced a fast entropy drop in the early stage, reflecting the transition of the model from broad exploration to deterministic exploitation. The gradient norm curves (Figure~\ref{fig:batch_curve}(c)) show that the MUA-RL training process is stable and free from issues such as gradient explosion and divergence.

\paragraph{Rollout turns \& response length.} 
The number of rollout turns (Figure~\ref{fig:batch_curve}(e)) increases at the beginning of training and subsequently stabilizes at an average of about 21-23 turns, 
while the response length (Figure~\ref{fig:batch_curve}(f)) remains largely unchanged throughout the reinforcement learning process. This observation indicates that the improvements in model performance are not driven by producing longer responses, as in test-time scaling of reasoning models where performance scales with output length, but rather by leveraging more structured multi-turn interactions with the user and the database -- a finding consistent with the observations made by GLM-4.5~\cite{zeng2025glm}. 


\paragraph{Unique 4-gram Ratio.} The Unique 4-gram Ratio captures the lexical diversity of generated responses by measuring the proportion of distinct four-token sequences. As illustrated in Figure~\ref{fig:batch_curve}(g), MUA-RL-8B maintains relatively higher diversity at the early training stage, reflecting its reliance on varied linguistic expressions when interacting with the user. In contrast, MUA-RL-32B exhibits lower lexical diversity with a consistently declining trend, indicating that larger models tend to accomplish tasks by enhancing their tool-using capabilities rather than relying on surface-level linguistic variation.

\paragraph{All Correct Query Ratio \& All Wrong Query Ratio.} The All Correct Query Ratio measures the proportion of tasks whose rollouts are all correct, serving as a strict indicator of correctness. Conversely, the All Wrong Query Ratio quantifies the proportion of tasks with all rollouts being incorrect, reflecting complete failure cases. As training progresses, we observe a clear upward trend in the All Correct Query Ratio across all model scales. This result indicates that MUA-RL effectively enables models from succeeding occasionally to consistently generating correct rollouts. In contrast, the All Wrong Query Ratio exhibits a steady decline during training, demonstrating the framework’s ability to suppress catastrophic failure cases. In Appendix~\ref{app:case study}, we present a case where, after applying MUA-RL, the outcomes improved from all rollouts being wrong to all rollouts being correct.

\begin{figure}[htbp]
  \centering
  \includegraphics[width=1.0\textwidth]{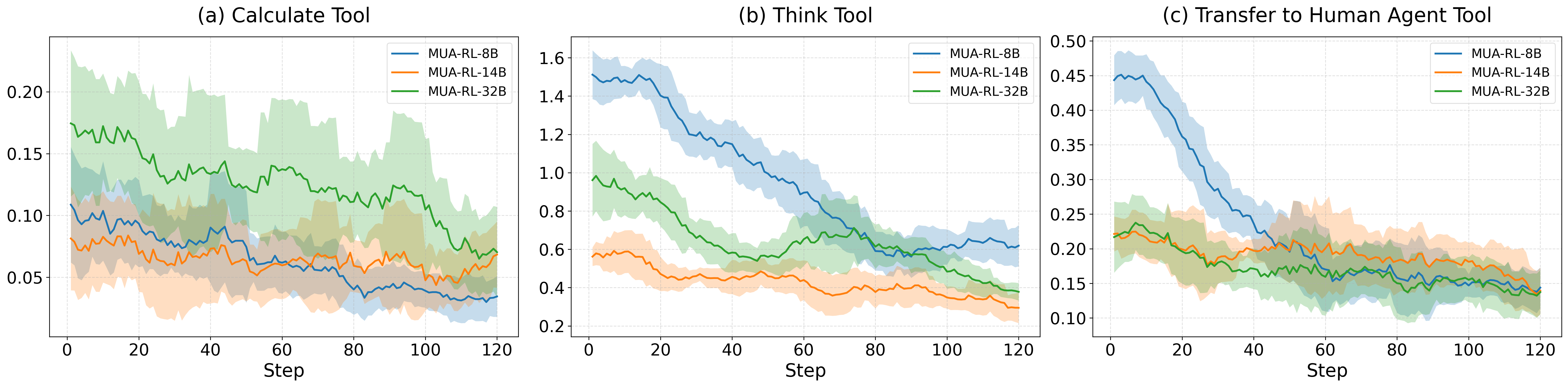}
  \caption{Trend in the average of general-purpose tools' invocation times during RL training.}
  \label{fig:tool_curve}
\end{figure}

\paragraph{General-purpose tools' invocation times.} In Figure~\ref{fig:tool_curve}, we present the trends in the invocation frequency of three general-purpose tools: Calculate, Think, and Transfer to Human Agent. Here, the Calculate tool performs simple mathematical computations, the Think tool enables non-thinking models to possess a certain level of reasoning capability, and the Transfer to Human Agent is called when the model deems itself unable to complete a task autonomously and transfers it to a human. As can be observed, the usage frequency of all three tools shows a declining trend. The decline could be attributed to the fact that MUA-RL effectively reduces reliance on tools with limited contribution to task completion, thereby improving both robustness and efficiency in real scenarios. For example, the model's reduction in calls to the Think tool results in a shorter and more efficient decision-making path.


\subsubsection{Ablation Study}

To evaluate the contribution of the cold-start stage in the MUA-RL framework, we perform an ablation study by applying reinforcement learning directly to Qwen3-32B Non-thinking without cold-start. We compare three models: (1) MUA-RL-32B, (2) MUA-RL-32B \textit{w/o} RL, and (3) MUA-RL-32B \textit{w/o} cold-start. The results are reported in Table~\ref{tab:ablation-32b}.

As can be observed from the table, MUA-RL-32B w/o cold-start achieves higher scores on TAU2-Bench compared to MUA-RL-32B w/o RL; however, it performs worse on BFCL-V3 Multi Turn. Nevertheless, both variants underperform relative to MUA-RL-32B on both TAU2-Bench and BFCL-V3 Multi Turn. This indicates that eliminating any stage leads to notable performance degradation, underscoring the necessity of every stage in the overall framework. The full pipeline, which integrates both cold-start and reinforcement learning, consistently delivers superior performance across all benchmarks.

\begin{table}[htbp]
\centering
\small
\resizebox{\textwidth}{!}{
\begin{tabular}{l ccc ccccc}
\toprule
    & \multicolumn{3}{c}{\textbf{TAU2}} 
    & \multicolumn{5}{c}{\textbf{BFCL-V3 Multi Turn}} \\
\cmidrule(lr){2-4} \cmidrule(lr){5-9}
\textbf{Model}
    & \textit{Retail}
    & \textit{Airline}
    & \textit{Telecom}
    & \textit{Base}
    & \makecell[c]{\textit{Miss}\\\textit{Func}}
    & \makecell[c]{\textit{Miss}\\\textit{Param}}
    & \makecell[c]{\textit{Long}\\\textit{Context}}
    & \makecell[c]{\textit{Overall}\\\textit{Acc}} \\
\midrule
Qwen3-32B Non-thinking
    & 50.2 
    & 23.5 
    & 24.8
    & 29.5
    & 11.0
    & 20.0
    & 18.0
    & 19.6 \\
\ \ + MUA-RL \textit{w/o} RL
    & 58.2\textsubscript{(\textcolor{teal}{↑8.0}, \textcolor{red}{↓9.1})} 
    & 31.1\textsubscript{(\textcolor{teal}{↑7.6}, \textcolor{red}{↓14.3})} 
    & 19.3\textsubscript{(\textcolor{red}{↓5.5}, \textcolor{red}{↓9.0})} 
    & 35.0\textsubscript{(\textcolor{teal}{↑5.5}, \textcolor{red}{↓7.0})} 
    & \textbf{21.0}\textsubscript{(\textcolor{teal}{↑10.0}, \textcolor{teal}{↑1.0})}  
    & 28.5\textsubscript{(\textcolor{teal}{↑8.5}, \textcolor{red}{↓1.5})} 
    & 19.5\textsubscript{(\textcolor{teal}{↑1.5}, \textcolor{red}{↓2.0})} 
    & 26.0\textsubscript{(\textcolor{teal}{↑6.4}, \textcolor{red}{↓2.4})} \\
\ \ + MUA-RL \textit{w/o} cold-start
    & 61.6\textsubscript{(\textcolor{teal}{↑11.4}, \textcolor{red}{↓5.7})} 
    & 41.0\textsubscript{(\textcolor{teal}{↑17.5}, \textcolor{red}{↓4.4})} 
    & 28.1\textsubscript{(\textcolor{teal}{↑3.3}, \textcolor{red}{↓0.2})} 
    & 29.5\textsubscript{(\textcolor{black}{0.0}, \textcolor{red}{↓12.5})} 
    & 13.0\textsubscript{(\textcolor{teal}{↑2.0}, \textcolor{red}{↓7.0})} 
    & 20.0\textsubscript{(\textcolor{black}{0.0}, \textcolor{red}{↓10.0})} 
    & 14.5\textsubscript{(\textcolor{red}{↓3.5}, \textcolor{red}{↓7.0})} 
    & 19.3\textsubscript{(\textcolor{red}{↓0.3}, \textcolor{red}{↓9.1})} \\
\midrule
MUA-RL-32B
    & \textbf{67.3} & \textbf{45.4} & \textbf{28.3} 
    & \textbf{42.0} & 20.0 & \textbf{30.0} & \textbf{21.5} & \textbf{28.4} \\
\bottomrule
\end{tabular}
}
\vspace{4pt}
\caption{Ablation study on MUA-RL-32B. Each model runs in non-thinking mode. 
\textit{w/o} denotes removing specific components during training. Each row shows performance changes in parentheses, 
where the first value indicates the change relative to the Qwen3-32B Non-thinking baseline, and the second indicates the change relative to the MUA-RL-32B. Performance gains are highlighted in \textcolor{teal}{green}, while drops are marked in \textcolor{red}{red}.}
\label{tab:ablation-32b}
\end{table}

\section{Conclusion}

In this work, we proposed MUA-RL, a reinforcement learning framework for multi-turn user-interacting agentic tool use. By combining a lightweight cold-start phase with Group Relative Policy Optimization (GRPO) on real-time interactive rollouts, MUA-RL effectively addresses the challenges of dynamic, user-driven interactions. Our design of simplified yet task-oriented reward enables models to autonomously develop robust behavioral patterns without overfitting to specific trajectories. Extensive evaluations on TAU1-Bench, TAU2-Bench, BFCL-V3 Multi Turn, and ACEBench Agent demonstrate that MUA-RL consistently outperforms its base and cold-start counterparts, and in several cases achieves performance comparable to or exceeding much larger models such as GPT-4o and DeepSeek-V3-0324. Notably, MUA-RL exhibits strong generalization in challenging domains like telecom dual-control tasks, confirming its adaptability to complex real-world scenarios. Further analyses of training dynamics, ablation studies, and tool-usage behaviors validate the stability and effectiveness of the framework.

Overall, MUA-RL provides a scalable, efficient, and generalizable reinforcement learning paradigm for agentic tool use in multi-turn user-interacting settings. We believe this work opens new opportunities for building more reliable and interactive LLM-based agents. Future research may extend this framework to richer multi-modal environments, explore more fine-grained reward shaping strategies, and further enhance alignment with human preferences in open-ended interactive tasks.

\section*{Acknowledgments}
We are grateful to the VeRL team for their infrastructure support. We would like to appreciate insightful discussions with Haoxing Wen, Peng Zhao.

\bibliographystyle{unsrt}  
\bibliography{references}

\newpage

\begin{appendices}
\section{Cold-start Data}
\label{appen:csd}




\begin{tcolorbox}[
    colback=gray!10, 
    colframe=black, 
    width=\textwidth, 
    boxrule=0.5pt,
    breakable,           
    enhanced,            
    before skip=10pt,    
    after skip=10pt,     
    pad at break*=1mm,   
    left=2mm,                 
    right=2mm,
    parbox=false,            
]

\textbf{Example \#1 LLM-simulated Tool Responses}

\vspace{0.3em}
\hrule height 0.5pt
\vspace{0.5em}

\textbf{SYSTEM PROMPT:}
\# University Course Registration Agent Policy

As a university course registration agent, you are responsible for helping students enroll in, drop, swap, or audit courses, as well as update their personal and academic information.

- You must authenticate the student at the start of every conversation by requesting their student id and verifying it with their full name and date of birth.

- You may only take actions affecting the authenticated student. Requests for actions regarding another student must be denied.

- For any action that changes the enrollment database (enrollment, dropping, swapping, auditing courses, or updating personal/academic information), you must clearly summarize the requested change and obtain explicit confirmation ('yes') from the student before proceeding.

- Only one action may be taken at a time. If you need to use a database tool, do so separately from messaging the student.

- Do not offer opinions, make recommendations, or generate information not available from the student or university records.

- Transfer the student to a human advisor only if the request falls outside the scope of your allowed actions or if there is a system error.

\#\# Domain Specifics

- All time references are in the university's local time (EST).

- Student records contain: student id, full name, date of birth, declared major(s)/minor(s), current academic year, academic standing, address, and a list of registered courses with associated grades.

- Courses have: course code, course title, department, semester offered, section number, instructor, meeting times, and a maximum capacity. Some courses have prerequisites or require departmental approval.

\#\# Registering for Courses

- Registration is allowed only during the official registration period for the upcoming semester. You must check the current date against registration period dates.

- Before enrolling, verify that:
  - The course is open for enrollment and not full.
  - The student meets all prerequisites.
  - There is no scheduling conflict with the student's existing courses.
  - The student does not exceed the maximum allowed course load (typically 5 courses per semester).
  - If the course requires departmental approval, advise the student to seek approval before proceeding.

- Upon successful registration, update the student's course list and confirm with the student.

\#\# Dropping or Withdrawing from Courses

- Dropping a course is permitted until the add/drop deadline; withdrawing is permitted until the withdrawal deadline (both set by the academic calendar).

- Before dropping/withdrawing, confirm the course code and reason for the request (schedule conflict, overloading, personal reason, etc.).

- Remind the student of any academic or financial implications.

\#\# Swapping Courses

- Swapping is only allowed during the registration period and if both courses involved have open seats and no scheduling conflict.

- Swapping is a one-step action (drop one, add another); both actions must succeed or neither is completed.

\#\# Auditing Courses

- Auditing is available only for students in good academic standing and if the course allows auditing.

- Audited courses do not count towards course load or GPA.

- You must confirm student understanding before proceeding.

\#\# Updating Personal/Academic Information

- Students may update address, phone number, and declared major/minor, but not academic history or grades.

- Requests to change declared major/minor must be checked for eligibility (GPA 3.5+).

- Changes take effect immediately for address/phone; major/minor changes require confirmation of eligibility.

- You must remind students to consult academic advisors for any major academic changes or concerns. 
\vspace{0.3em}
\hrule height 0.5pt
\vspace{0.5em}

\textbf{Available Tools:}

\texttt{authenticate\_student}, \texttt{get\_student\_record}, \texttt{get\_course\_details}, \texttt{list\_available\_courses}, \texttt{check\_registration\_period}, \texttt{register\_for\_course}, \texttt{drop\_course}, \texttt{withdraw\_from\_course}, \texttt{swap\_courses}, \texttt{audit\_course}, \texttt{update\_address\_or\_phone}, \texttt{update\_major\_or\_minor}, \texttt{think}, \texttt{transfer\_to\_human\_advisor}

\vspace{0.3em}
\hrule height 0.5pt
\vspace{0.5em}

\textbf{Trajectory:}

\textbf{user}:Hi, I need to update my contact details with the university.

\textbf{assistant}:Hello! I'd be happy to help you update your contact details with the university. Before we proceed, I need to authenticate your identity for security purposes.

Could you please provide me with the following information:
- Your student ID
- Your full legal name
- Your date of birth (in YYYY-MM-DD format)

Once I verify this information, I'll be able to help you update your address and/or phone number in the university system.

\textbf{user}:Sure, my student ID is S32165498, my name is Ethan Williams, and my date of birth is 2004-12-03.

\textbf{assistant}:\{'name': 'authenticate\_student', 'arguments': '\{"student\_id":"S32165498","full\_name":"Ethan Williams","date\_of\_birth":"2004-12-03"\}'\}

\textbf{tool}:\{"authenticated":true,"student\_id":"S32165498"\}

\textbf{assistant}:Thank you for providing your information, Ethan. Your identity has been successfully authenticated.

I can now help you update your contact details. What specific information would you like to update? You can update your:
- Address
- Phone number
- Or both

Please let me know what changes you'd like to make.

\textbf{user}:I just need to update my phone number, please. My address should stay the same.

\textbf{assistant}:I'd be happy to help you update your phone number. Could you please provide your new phone number that you'd like to have on record?

\textbf{user}:My new phone number is 555-123-4321.

\textbf{assistant}:Thank you for providing your new phone number. Before I make this change, I want to confirm:

You would like to update your phone number to: 555-123-4321
Your address will remain unchanged.

Is this correct? Please confirm with a "yes" if you'd like me to proceed with this update.

\textbf{user}:Yes, that's correct. Please go ahead.

\textbf{assistant}:\{'name': 'update\_address\_or\_phone', 'arguments': '\{"student\_id":"S32165498","phone":"555-123-4321","address":""\}'\}

\textbf{tool}:\{"success":true,"student\_id":"S32165498","address":"","phone":"555-123-4321","message":"Address/phone updated."\}

\textbf{assistant}:Great! I've successfully updated your phone number to 555-123-4321 in the university system. Your address remains unchanged as requested.

The update has been processed immediately and is now reflected in your student record. Is there anything else you need help with today?

\textbf{user}:\#\#\#STOP\#\#\#

\end{tcolorbox}

\begin{tcolorbox}[
    colback=gray!10, 
    colframe=black, 
    width=\textwidth, 
    boxrule=0.5pt,
    breakable,           
    enhanced,            
    before skip=10pt,    
    after skip=10pt,     
    pad at break*=1mm,   
    left=2mm,                 
    right=2mm,
    parbox=false,            
]

\textbf{Example \#2 MCP Server Tool Responses}

\vspace{0.3em}
\hrule height 0.5pt
\vspace{0.5em}

\textbf{SYSTEM PROMPT:}

\# AniList Agent Policy

The current time is 2025-05-27 15:00:00 CST (Beijing, China).

As an AniList agent, you can help users search for, retrieve, and filter information on anime, manga, studios, staff, characters, users, activities, and site statistics.The user is assumed to be already logged in by default.

\#\# General Guidelines

- You must only use one tool call at a time. If you make a tool call, do not respond to the user in the same step. If you respond to the user, do not make a tool call at the same time.
- You must not provide information, knowledge, or recommendations not directly retrieved from the tools or provided by the user. Do not speculate, invent, or comment subjectively.
- You must transfer the user to a human agent if, and only if, the user explicitly requests a human, or the request cannot be fulfilled with the available tools.
- Deny any request that is outside the capabilities of your tools or against this policy.
- AniList distinguishes between IDs and names for entities such as users, studios, staff, and characters. Always use the correct type as required by the tool.
- Do not attempt to modify or update the database or user state by any means other than the allowed tools.

\#\# Viewing, Searching, and Retrieving Information

- When searching for anime, manga, characters, staff, or studios, if the user provides ambiguous or broad criteria (e.g., a general term that could be a genre, tag, or title), first use 'get\_genres' and 'get\_media\_tags' to clarify intent before issuing search queries. Only use query terms for actual title or name searches; otherwise, use filters.
- You should not make up filter values for genres, tags, or media status. Always use values from 'get\_genres' and 'get\_media\_tags' where applicable.
- When paginating search or list results, always respect the 'amount'/'perPage' and 'page' parameters and do not attempt to retrieve more than the allowed maximum (typically 25 per request).

\#\# User Profile and Account Actions

- You must only perform favorite/unfavorite or follow/unfollow actions, for the currently logged-in user and never for another user. Requests to modify another user's data must be denied.
- For any tool that affects the user's AniList account (such as favouriting/unfavouriting or following/unfollowing users), you must:
    - Explicitly confirm with the user before performing the action, summarizing what will happen.
    - Proceed only after the user replies with explicit confirmation (e.g., "yes").

\vspace{0.3em}
\hrule height 0.5pt
\vspace{0.5em}

\textbf{Available Tools:}

\texttt{get\_genres}, \texttt{get\_media\_tags}, \texttt{get\_site\_statistics}, \texttt{get\_studio}, \texttt{get\_activity}, \texttt{get\_user\_activity}, \texttt{get\_user\_anime\_list}, \texttt{get\_user\_manga\_list}, \texttt{get\_anime}, \texttt{favourite\_anime}, \texttt{favourite\_manga}, \texttt{get\_manga}, \texttt{get\_character}, \texttt{favourite\_character}, \texttt{favourite\_staff}, \texttt{get\_staff}, \texttt{get\_recommendation}, \texttt{get\_recommendations\_for\_media}, \texttt{search\_activity}, \texttt{search\_anime}, \texttt{search\_character}, \texttt{search\_manga}, \texttt{search\_staff}, \texttt{search\_studio}, \texttt{search\_user}, \texttt{get\_full\_user\_info}, \texttt{follow\_user}, \texttt{get\_authorized\_user}, \texttt{get\_user\_recent\_activity}, \texttt{get\_user\_profile}, \texttt{get\_user\_stats}, \texttt{think}, \texttt{transfer\_to\_human\_agents}

\vspace{0.3em}
\hrule height 0.5pt
\vspace{0.5em}

\textbf{Trajectory:}

\textbf{user}:Hi! Can you show me a list of characters whose name includes "Sakura"?

\textbf{assistant}:\{'name': 'search\_character', 'arguments': '\{"term":"Sakura"\}'\}

\textbf{tool}:\{
  "pageInfo": \{
    "total": 5000,
    "currentPage": 1,
    "lastPage": 1000,
    "hasNextPage": true,
    "perPage": 5
  \},
  "characters": [
    \{
      "id": 500,
      "name": \{
        "english": "Sakura Matou"
      \}
    \},
    \{
      "id": 127222,
      "name": \{
        "english": "Mai Sakurajima"
      \}
    \},
    \{
      "id": 127118,
      "name": \{
        "english": "Sakura Yamauchi"
      \}
    \},
    \{
      "id": 2671,
      "name": \{
        "english": "Sakura Kinomoto"
      \}
    \},
    \{
      "id": 145,
      "name": \{
        "english": "Sakura Haruno"
      \}
    \}
  ]
\}

\textbf{assistant}:Here are characters whose names include "Sakura":

1. **Sakura Matou** (ID: 500)
2. **Mai Sakurajima** (ID: 127222)
3. **Sakura Yamauchi** (ID: 127118)
4. **Sakura Kinomoto** (ID: 2671)
5. **Sakura Haruno** (ID: 145)

This shows the first 5 results out of 5,000 total characters found. The search found characters from various popular anime and manga series. Would you like me to show more results from the next page, or would you like detailed information about any specific character from this list?

\textbf{user}:Could you show me 15 characters instead of just 5?

\textbf{assistant}:\{'name': 'search\_character', 'arguments': '\{"term":"Sakura","amount":15\}'\}

\textbf{tool}:\{
  "pageInfo": \{
    "total": 5000,
    "currentPage": 1,
    "lastPage": 333,
    "hasNextPage": true,
    "perPage": 15
  \},
  "characters": [
    \{
      "id": 500,
      "name": \{
        "english": "Sakura Matou"
      \}
    \},
    \{
      "id": 127222,
      "name": \{
        "english": "Mai Sakurajima"
      \}
    \},
    \{
      "id": 127118,
      "name": \{
        "english": "Sakura Yamauchi"
      \}
    \},
    \{
      "id": 2671,
      "name": \{
        "english": "Sakura Kinomoto"
      \}
    \},
    \{
      "id": 145,
      "name": \{
        "english": "Sakura Haruno"
      \}
    \},
    \{
      "id": 40006,
      "name": \{
        "english": "Kyouko Sakura"
      \}
    \},
    \{
      "id": 87271,
      "name": \{
        "english": "Chiyo Sakura "
      \}
    \},
    \{
      "id": 230204,
      "name": \{
        "english": "Haruka Sakura"
      \}
    \},
    \{
      "id": 121635,
      "name": \{
        "english": "Futaba Sakura"
      \}
    \},
    \{
      "id": 310,
      "name": \{
        "english": "Hanamichi Sakuragi"
      \}
    \},
    \{
      "id": 144717,
      "name": \{
        "english": "Sakura Adachi"
      \}
    \},
    \{
      "id": 144665,
      "name": \{
        "english": "Sumi Sakurasawa"
      \}
    \},
    \{
      "id": 123215,
      "name": \{
        "english": "Airi Sakura"
      \}
    \},
    \{
      "id": 132856,
      "name": \{
        "english": "Hibiki Sakura"
      \}
    \},
    \{
      "id": 72449,
      "name": \{
        "english": "Sakura Kouno"
      \}
    \}
  ]
\}

\textbf{assistant}:Here are 15 characters whose names include "Sakura":

1. **Sakura Matou** (ID: 500)
2. **Mai Sakurajima** (ID: 127222)
3. **Sakura Yamauchi** (ID: 127118)
4. **Sakura Kinomoto** (ID: 2671)
5. **Sakura Haruno** (ID: 145)
6. **Kyouko Sakura** (ID: 40006)
7. **Chiyo Sakura** (ID: 87271)
8. **Haruka Sakura** (ID: 230204)
9. **Futaba Sakura** (ID: 121635)
10. **Hanamichi Sakuragi** (ID: 310)
11. **Sakura Adachi** (ID: 144717)
12. **Sumi Sakurasawa** (ID: 144665)
13. **Airi Sakura** (ID: 123215)
14. **Hibiki Sakura** (ID: 132856)
15. **Sakura Kouno** (ID: 72449)

This shows 15 results out of 5,000 total characters found with "Sakura" in their names. Would you like to see more results, or get detailed information about any specific character from this list?

\textbf{user}:That's perfect, thanks! I'm good for now. 

\#\#\#STOP\#\#\#

\end{tcolorbox}

\section{Case Study}
\label{app:case study}

A case study is conducted to illustrate the model's behaviors before and after MUA-RL. It includes three parts: SYSTEM PROMPT, Available Tools, and Trajectory. Specifically, the case study demonstrates how MUA-RL reshapes the model’s behavior in practice. As shown in the trajectory, without MUA-RL, the baseline model (Qwen3-32B Non-thinking) prematurely modifies the order without obtaining explicit confirmation from the user, which not only violates the policy but also leads to a failed modification when the user requests further change. By contrast, in the correct trajectory after MUA-RL, the model not only adheres strictly to the policy but also demonstrates a more natural and reliable interaction pattern. The model first authenticates the user's identity, then carefully lists all available product variants, clearly explains the price differences, and explicitly requests confirmation before executing the modification. This improvement highlights that MUA-RL does not make the model rigidly follow the policy, but directly transforms the model's interaction strategy into a cautious, policy-grounded, and user-centered one.

\begin{tcolorbox}[
    colback=gray!10, 
    colframe=black, 
    width=\textwidth, 
    boxrule=0.5pt,
    breakable,           
    enhanced,            
    before skip=10pt,    
    after skip=10pt,     
    pad at break*=1mm,   
    left=2mm,                 
    right=2mm,
    parbox=false,            
]

\textbf{Example from TAU2 Retail}

\vspace{0.3em}
\hrule height 0.5pt
\vspace{0.5em}

\textbf{SYSTEM PROMPT:}

\# Retail agent policy

As a retail agent, you can help users:

- **cancel or modify pending orders**
- **return or exchange delivered orders**
- **modify their default user address**
- **provide information about their own profile, orders, and related products**

At the beginning of the conversation, you have to authenticate the user identity by locating their user id via email, or via name + zip code. This has to be done even when the user already provides the user id.

Once the user has been authenticated, you can provide the user with information about order, product, profile information, e.g. help the user look up order id.

You can only help one user per conversation (but you can handle multiple requests from the same user), and must deny any requests for tasks related to any other user.

Before taking any action that updates the database (cancel, modify, return, exchange), you must list the action details and obtain explicit user confirmation (yes) to proceed.

You should not make up any information or knowledge or procedures not provided by the user or the tools, or give subjective recommendations or comments.

You should at most make one tool call at a time, and if you take a tool call, you should not respond to the user at the same time. If you respond to the user, you should not make a tool call at the same time.

You should deny user requests that are against this policy.

You should transfer the user to a human agent if and only if the request cannot be handled within the scope of your actions. To transfer, first make a tool call to transfer\_to\_human\_agents, and then send the message 'YOU ARE BEING TRANSFERRED TO A HUMAN AGENT. PLEASE HOLD ON.' to the user.

\#\# Domain basic

- All times in the database are EST and 24 hour based. For example "02:30:00" means 2:30 AM EST.

\#\#\# User

Each user has a profile containing:

- unique user id
- email
- default address
- payment methods.

There are three types of payment methods: **gift card**, **paypal account**, **credit card**.

\#\#\# Product

Our retail store has 50 types of products.

For each **type of product**, there are **variant items** of different **options**.

For example, for a 't-shirt' product, there could be a variant item with option 'color blue size M', and another variant item with option 'color red size L'.

Each product has the following attributes:

- unique product id
- name
- list of variants

Each variant item has the following attributes:

- unique item id
- information about the value of the product options for this item.
- availability
- price

Note: Product ID and Item ID have no relations and should not be confused!

\#\#\# Order

Each order has the following attributes:

- unique order id
- user id
- address
- items ordered
- status
- fulfillments info (tracking id and item ids)
- payment history

The status of an order can be: **pending**, **processed**, **delivered**, or **cancelled**.

Orders can have other optional attributes based on the actions that have been taken (cancellation reason, which items have been exchanged, what was the exchange price difference etc)

\#\# Generic action rules

Generally, you can only take action on pending or delivered orders.

Exchange or modify order tools can only be called once per order. Be sure that all items to be changed are collected into a list before making the tool call!!!

\#\# Cancel pending order

An order can only be cancelled if its status is 'pending', and you should check its status before taking the action.

The user needs to confirm the order id and the reason (either 'no longer needed' or 'ordered by mistake') for cancellation. Other reasons are not acceptable.

After user confirmation, the order status will be changed to 'cancelled', and the total will be refunded via the original payment method immediately if it is a gift card, otherwise in 5 to 7 business days.

\#\# Modify pending order

An order can only be modified if its status is 'pending', and you should check its status before taking the action.

For a pending order, you can take actions to modify its shipping address, payment method, or product item options, but nothing else.

\#\#\# Modify payment

The user can only choose a single payment method different from the original payment method.

If the user wants the modify the payment method to gift card, it must have enough balance to cover the total amount.

After user confirmation, the order status will be kept as 'pending'. The original payment method will be refunded immediately if it is a gift card, otherwise it will be refunded within 5 to 7 business days.

\#\#\# Modify items

This action can only be called once, and will change the order status to 'pending (items modified)'. The agent will not be able to modify or cancel the order anymore. So you must confirm all the details are correct and be cautious before taking this action. In particular, remember to remind the customer to confirm they have provided all the items they want to modify.

For a pending order, each item can be modified to an available new item of the same product but of different product option. There cannot be any change of product types, e.g. modify shirt to shoe.

The user must provide a payment method to pay or receive refund of the price difference. If the user provides a gift card, it must have enough balance to cover the price difference.

\#\# Return delivered order

An order can only be returned if its status is 'delivered', and you should check its status before taking the action.

The user needs to confirm the order id and the list of items to be returned.

The user needs to provide a payment method to receive the refund.

The refund must either go to the original payment method, or an existing gift card.

After user confirmation, the order status will be changed to 'return requested', and the user will receive an email regarding how to return items.

\#\# Exchange delivered order

An order can only be exchanged if its status is 'delivered', and you should check its status before taking the action. In particular, remember to remind the customer to confirm they have provided all items to be exchanged.

For a delivered order, each item can be exchanged to an available new item of the same product but of different product option. There cannot be any change of product types, e.g. modify shirt to shoe.

The user must provide a payment method to pay or receive refund of the price difference. If the user provides a gift card, it must have enough balance to cover the price difference.

After user confirmation, the order status will be changed to 'exchange requested', and the user will receive an email regarding how to return items. There is no need to place a new order.

\vspace{0.3em}
\hrule height 0.5pt
\vspace{0.5em}

\textbf{Available Tools:}

\texttt{calculate}, \texttt{cancel\_pending\_order}, \texttt{exchange\_delivered\_order\_items}, \texttt{find\_user\_id\_by\_name\_zip}, \texttt{find\_user\_id\_by\_email}, \texttt{get\_order\_details}, \texttt{get\_product\_details}, \texttt{get\_user\_details}, \texttt{list\_all\_product\_types}, \texttt{modify\_pending\_order\_address}, \texttt{modify\_pending\_order\_items}, \texttt{modify\_pending\_order\_payment}, \texttt{modify\_user\_address}, \texttt{return\_delivered\_order\_items}, \texttt{transfer\_to\_human\_agents}

\vspace{0.3em}
\hrule height 0.5pt
\vspace{0.5em}

\textcolor{red}{Error Trajectory in Qwen3-32B Non-thinking:}

\textbf{assistant}:Hi! How can I help you today?

\textbf{user}:Hi, I’d like to change the wireless earbuds in my order W5061109 to a blue colored version. My name is Chen Johnson and I’m in Houston, TX, 77004. Can you please make sure the price stays the same or is lower, and let me know explicitly? I’d appreciate it if you could resolve this as soon as possible. Thanks!

\textbf{assistant}:\{'name': 'find\_user\_id\_by\_email', 'arguments': \{'email': 'chen.johnson@example.com'\}\}

\textbf{tool}:Error: User not found

\textbf{assistant}:\{'name': 'find\_user\_id\_by\_name\_zip', 'arguments': \{'first\_name': 'Chen', 'last\_name': 'Johnson', 'zip': '77004'\}\}

\textbf{tool}:chen\_johnson\_4204

\textbf{assistant}:\{'name': 'get\_order\_details', 'arguments': \{'order\_id': '\#W5061109'\}\}

\textbf{tool}:\{"order\_id": "\#W5061109", "user\_id": "chen\_johnson\_4204", "address": \{"address1": "503 Elm Avenue", "address2": "Suite 641", "city": "Houston", "country": "USA", "state": "TX", "zip": "77004"\}, "items": [\{"name": "Makeup Kit", "product\_id": "5149340237", "item\_id": "6254646215", "price": 248.85, "options": \{"skin tone": "dark", "kit size": "basic", "brand": "Brand B"\}\}, \{"name": "Wireless Earbuds", "product\_id": "9924732112", "item\_id": "3694871183", "price": 256.67, "options": \{"color": "white", "battery life": "8 hours", "water resistance": "IPX4"\}\}, \{"name": "Office Chair", "product\_id": "4794339885", "item\_id": "8323284863", "price": 511.24, "options": \{"material": "fabric", "color": "blue", "armrest": "adjustable", "backrest height": "standard"\}\}, \{"name": "Bluetooth Speaker", "product\_id": "4768869376", "item\_id": "3254583681", "price": 302.67, "options": \{"color": "blue", "battery life": "20 hours", "water resistance": "yes"\}\}], "status": "pending", "fulfillments": [], "payment\_history": [\{"transaction\_type": "payment", "amount": 1319.43, "payment\_method\_id": "paypal\_3742148"\}], "cancel\_reason": null, "exchange\_items": null, "exchange\_new\_items": null, "exchange\_payment\_method\_id": null, "exchange\_price\_difference": null, "return\_items": null, "return\_payment\_method\_id": null\}

\textbf{assistant}:\{'name': 'get\_product\_details', 'arguments': \{'product\_id': '9924732112'\}\}

\textbf{tool}:\{"name": "Wireless Earbuds", "product\_id": "9924732112", "variants": \{"9580569596": \{"item\_id": "9580569596", "options": \{"color": "black", "battery life": "4 hours", "water resistance": "IPX7"\}, "available": true, "price": 257.38\}, "2499294441": \{"item\_id": "2499294441", "options": \{"color": "black", "battery life": "8 hours", "water resistance": "IPX7"\}, "available": false, "price": 258.36\}, "1646531091": \{"item\_id": "1646531091", "options": \{"color": "blue", "battery life": "6 hours", "water resistance": "IPX4"\}, "available": true, "price": 232.49\}, "8555936349": \{"item\_id": "8555936349", "options": \{"color": "blue", "battery life": "8 hours", "water resistance": "IPX4"\}, "available": true, "price": 226.49\}, "5565631513": \{"item\_id": "5565631513", "options": \{"color": "black", "battery life": "6 hours", "water resistance": "IPX7"\}, "available": false, "price": 267.9\}, "6077640618": \{"item\_id": "6077640618", "options": \{"color": "blue", "battery life": "8 hours", "water resistance": "not resistant"\}, "available": true, "price": 242.92\}, "9270970345": \{"item\_id": "9270970345", "options": \{"color": "black", "battery life": "6 hours", "water resistance": "not resistant"\}, "available": false, "price": 259.03\}, "4063058357": \{"item\_id": "4063058357", "options": \{"color": "black", "battery life": "4 hours", "water resistance": "not resistant"\}, "available": true, "price": 243.34\}, "3694871183": \{"item\_id": "3694871183", "options": \{"color": "white", "battery life": "8 hours", "water resistance": "IPX4"\}, "available": false, "price": 256.67\}, "6452271382": \{"item\_id": "6452271382", "options": \{"color": "blue", "battery life": "4 hours", "water resistance": "IPX4"\}, "available": true, "price": 258.84\}, "2052249669": \{"item\_id": "2052249669", "options": \{"color": "white", "battery life": "4 hours", "water resistance": "not resistant"\}, "available": true, "price": 237.14\}, "2757705742": \{"item\_id": "2757705742", "options": \{"color": "blue", "battery life": "4 hours", "water resistance": "IPX7"\}, "available": false, "price": 258.97\}\}\}

\textbf{assistant}:\{'name': 'modify\_pending\_order\_items', 'arguments': \{'order\_id': '\#W5061109', 'item\_ids': ['3694871183'], 'new\_item\_ids': ['8555936349'], 'payment\_method\_id': 'paypal\_3742148'\}\}

\textbf{tool}:\{"order\_id": "\#W5061109", "user\_id": "chen\_johnson\_4204", "address": \{"address1": "503 Elm Avenue", "address2": "Suite 641", "city": "Houston", "country": "USA", "state": "TX", "zip": "77004"\}, "items": [\{"name": "Makeup Kit", "product\_id": "5149340237", "item\_id": "6254646215", "price": 248.85, "options": \{"skin tone": "dark", "kit size": "basic", "brand": "Brand B"\}\}, \{"name": "Wireless Earbuds", "product\_id": "9924732112", "item\_id": "8555936349", "price": 226.49, "options": \{"color": "blue", "battery life": "8 hours", "water resistance": "IPX4"\}\}, \{"name": "Office Chair", "product\_id": "4794339885", "item\_id": "8323284863", "price": 511.24, "options": \{"material": "fabric", "color": "blue", "armrest": "adjustable", "backrest height": "standard"\}\}, \{"name": "Bluetooth Speaker", "product\_id": "4768869376", "item\_id": "3254583681", "price": 302.67, "options": \{"color": "blue", "battery life": "20 hours", "water resistance": "yes"\}\}], "status": "pending (item modified)", "fulfillments": [], "payment\_history": [\{"transaction\_type": "payment", "amount": 1319.43, "payment\_method\_id": "paypal\_3742148"\}, \{"transaction\_type": "refund", "amount": 30.180000000000007, "payment\_method\_id": "paypal\_3742148"\}], "cancel\_reason": null, "exchange\_items": null, "exchange\_new\_items": null, "exchange\_payment\_method\_id": null, "exchange\_price\_difference": null, "return\_items": null, "return\_payment\_method\_id": null\}

\textbf{assistant}:The wireless earbuds in your order have been successfully changed to the blue-colored version with an 8-hour battery life and IPX4 water resistance. The price has been adjusted accordingly, and the total cost of your order has been updated. Let me know if there's anything else I can assist you with!

\textbf{user}:Thanks for the update! Could you please let me know if there’s a blue option without water resistance, and if so, can you switch to that one instead? I prefer the version without water resistance. Also, please confirm the price for that option.

\textbf{assistant}:\{'name': 'get\_product\_details', 'arguments': \{'product\_id': '9924732112'\}\}

\textbf{tool}:\{"name": "Wireless Earbuds", "product\_id": "9924732112", "variants": \{"9580569596": \{"item\_id": "9580569596", "options": \{"color": "black", "battery life": "4 hours", "water resistance": "IPX7"\}, "available": true, "price": 257.38\}, "2499294441": \{"item\_id": "2499294441", "options": \{"color": "black", "battery life": "8 hours", "water resistance": "IPX7"\}, "available": false, "price": 258.36\}, "1646531091": \{"item\_id": "1646531091", "options": \{"color": "blue", "battery life": "6 hours", "water resistance": "IPX4"\}, "available": true, "price": 232.49\}, "8555936349": \{"item\_id": "8555936349", "options": \{"color": "blue", "battery life": "8 hours", "water resistance": "IPX4"\}, "available": true, "price": 226.49\}, "5565631513": \{"item\_id": "5565631513", "options": \{"color": "black", "battery life": "6 hours", "water resistance": "IPX7"\}, "available": false, "price": 267.9\}, "6077640618": \{"item\_id": "6077640618", "options": \{"color": "blue", "battery life": "8 hours", "water resistance": "not resistant"\}, "available": true, "price": 242.92\}, "9270970345": \{"item\_id": "9270970345", "options": \{"color": "black", "battery life": "6 hours", "water resistance": "not resistant"\}, "available": false, "price": 259.03\}, "4063058357": \{"item\_id": "4063058357", "options": \{"color": "black", "battery life": "4 hours", "water resistance": "not resistant"\}, "available": true, "price": 243.34\}, "3694871183": \{"item\_id": "3694871183", "options": \{"color": "white", "battery life": "8 hours", "water resistance": "IPX4"\}, "available": false, "price": 256.67\}, "6452271382": \{"item\_id": "6452271382", "options": \{"color": "blue", "battery life": "4 hours", "water resistance": "IPX4"\}, "available": true, "price": 258.84\}, "2052249669": \{"item\_id": "2052249669", "options": \{"color": "white", "battery life": "4 hours", "water resistance": "not resistant"\}, "available": true, "price": 237.14\}, "2757705742": \{"item\_id": "2757705742", "options": \{"color": "blue", "battery life": "4 hours", "water resistance": "IPX7"\}, "available": false, "price": 258.97\}\}\}

\textbf{assistant}:The blue wireless earbuds without water resistance are available. The item ID is **6077640618**, and the price for this option is **\$242.92**. 

Would you like me to switch your order to this version?

\textbf{user}:Yes, please switch my order to the blue wireless earbuds without water resistance (item ID 6077640618) for \$242.92. Can you confirm that the price is the same or lower than my original order before making the change? Thank you!

\textbf{assistant}:\{'name': 'modify\_pending\_order\_items', 'arguments': \{'order\_id': '\#W5061109', 'item\_ids': ['8555936349'], 'new\_item\_ids': ['6077640618'], 'payment\_method\_id': 'paypal\_3742148'\}\}

\textbf{tool}:Error: Non-pending order cannot be modified

\textbf{assistant}:It seems there was an issue with modifying the order because it is no longer in a pending state. I will transfer your request to a human agent to assist you further. 

\textbf{assistant}:\{'name': 'transfer\_to\_human\_agents', 'arguments': \{'summary': 'User wants to modify a non-pending order to switch wireless earbuds to a blue version without water resistance.'\}\}

\textbf{tool}:Transfer successful

\textbf{assistant}:YOU ARE BEING TRANSFERRED TO A HUMAN AGENT. PLEASE HOLD ON.

\textbf{user}:\#\#\#TRANSFER\#\#\#

\vspace{0.3em}
\hrule height 0.5pt
\vspace{0.5em}

\textcolor{teal}{Correct Trajectory after MUA-RL:}

\textbf{assistant}:Hi! How can I help you today?

\textbf{user}:Hi, I’d like to change the wireless earbuds in my order W5061109 to a blue colored one. My name is Chen Johnson and I’m in Houston, TX, 77004. Can you please make sure the price stays the same or is lower, and let me know explicitly? I’d appreciate it if you could resolve this as soon as possible. Thanks!

\textbf{assistant}:I'll help you with changing the wireless earbuds in your order. First, I need to authenticate your account. You've provided your name and zip code, so I'll use that to find your user ID.

\textbf{assistant}:\{'name': 'find\_user\_id\_by\_name\_zip', 'arguments': \{'first\_name': 'Chen', 'last\_name': 'Johnson', 'zip': '77004'\}\}

\textbf{tool}:chen\_johnson\_4204

\textbf{assistant}:\{'name': 'get\_order\_details', 'arguments': \{'order\_id': '\#W5061109'\}\}

\textbf{tool}:\{"order\_id": "\#W5061109", "user\_id": "chen\_johnson\_4204", "address": \{"address1": "503 Elm Avenue", "address2": "Suite 641", "city": "Houston", "country": "USA", "state": "TX", "zip": "77004"\}, "items": [\{"name": "Makeup Kit", "product\_id": "5149340237", "item\_id": "6254646215", "price": 248.85, "options": \{"skin tone": "dark", "kit size": "basic", "brand": "Brand B"\}\}, \{"name": "Wireless Earbuds", "product\_id": "9924732112", "item\_id": "3694871183", "price": 256.67, "options": \{"color": "white", "battery life": "8 hours", "water resistance": "IPX4"\}\}, \{"name": "Office Chair", "product\_id": "4794339885", "item\_id": "8323284863", "price": 511.24, "options": \{"material": "fabric", "color": "blue", "armrest": "adjustable", "backrest height": "standard"\}\}, \{"name": "Bluetooth Speaker", "product\_id": "4768869376", "item\_id": "3254583681", "price": 302.67, "options": \{"color": "blue", "battery life": "20 hours", "water resistance": "yes"\}\}], "status": "pending", "fulfillments": [], "payment\_history": [\{"transaction\_type": "payment", "amount": 1319.43, "payment\_method\_id": "paypal\_3742148"\}], "cancel\_reason": null, "exchange\_items": null, "exchange\_new\_items": null, "exchange\_payment\_method\_id": null, "exchange\_price\_difference": null, "return\_items": null, "return\_payment\_method\_id": null\}

\textbf{assistant}:\{'name': 'get\_product\_details', 'arguments': \{'product\_id': '9924732112'\}\}

\textbf{tool}:\{"name": "Wireless Earbuds", "product\_id": "9924732112", "variants": \{"9580569596": \{"item\_id": "9580569596", "options": \{"color": "black", "battery life": "4 hours", "water resistance": "IPX7"\}, "available": true, "price": 257.38\}, "2499294441": \{"item\_id": "2499294441", "options": \{"color": "black", "battery life": "8 hours", "water resistance": "IPX7"\}, "available": false, "price": 258.36\}, "1646531091": \{"item\_id": "1646531091", "options": \{"color": "blue", "battery life": "6 hours", "water resistance": "IPX4"\}, "available": true, "price": 232.49\}, "8555936349": \{"item\_id": "8555936349", "options": \{"color": "blue", "battery life": "8 hours", "water resistance": "IPX4"\}, "available": true, "price": 226.49\}, "5565631513": \{"item\_id": "5565631513", "options": \{"color": "black", "battery life": "6 hours", "water resistance": "IPX7"\}, "available": false, "price": 267.9\}, "6077640618": \{"item\_id": "6077640618", "options": \{"color": "blue", "battery life": "8 hours", "water resistance": "not resistant"\}, "available": true, "price": 242.92\}, "9270970345": \{"item\_id": "9270970345", "options": \{"color": "black", "battery life": "6 hours", "water resistance": "not resistant"\}, "available": false, "price": 259.03\}, "4063058357": \{"item\_id": "4063058357", "options": \{"color": "black", "battery life": "4 hours", "water resistance": "not resistant"\}, "available": true, "price": 243.34\}, "3694871183": \{"item\_id": "3694871183", "options": \{"color": "white", "battery life": "8 hours", "water resistance": "IPX4"\}, "available": false, "price": 256.67\}, "6452271382": \{"item\_id": "6452271382", "options": \{"color": "blue", "battery life": "4 hours", "water resistance": "IPX4"\}, "available": true, "price": 258.84\}, "2052249669": \{"item\_id": "2052249669", "options": \{"color": "white", "battery life": "4 hours", "water resistance": "not resistant"\}, "available": true, "price": 237.14\}, "2757705742": \{"item\_id": "2757705742", "options": \{"color": "blue", "battery life": "4 hours", "water resistance": "IPX7"\}, "available": false, "price": 258.97\}\}\}

\textbf{assistant}:I can see that you have an order (\#W5061109) with wireless earbuds that you'd like to change to a blue color. Let me check the available options:

There are three blue-colored wireless earbuds available:
1. Item ID: 1646531091 - Blue, 6 hours battery life, IPX4 water resistance - \$232.49
2. Item ID: 8555936349 - Blue, 8 hours battery life, IPX4 water resistance - \$226.49
3. Item ID: 6077640618 - Blue, 8 hours battery life, not water resistant - \$242.92

Your current order has white earbuds (Item ID: 3694871183) priced at \$256.67. 

If you'd like to change to any of these blue options, I can help you modify your order. The price would be lower for options 1 and 2, and slightly higher for option 3. 

Please let me know which blue option you prefer, and I'll proceed with the modification after your confirmation.

\textbf{user}:Thanks for checking! I’d like to go with option 3—the blue earbuds (Item ID: 6077640618) that are not water resistant. Please confirm the final price for this change and let me know if there’s anything else you need from me to proceed.

\textbf{assistant}:I'll help you modify your order to change the wireless earbuds to the blue ones (Item ID: 6077640618). 

Current price of white earbuds: \$256.67
New price of blue earbuds: \$242.92
Price difference: -\$13.75 (you'll receive a refund of \$13.75)

Since this is a pending order, I can modify the item for you. Before proceeding, I need your explicit confirmation:

Would you like me to modify your order (\#W5061109) to change the white wireless earbuds (Item ID: 3694871183) to blue wireless earbuds (Item ID: 6077640618)? The price will decrease by \$13.75, and the refund will be processed to your original payment method (PayPal).

Please confirm with a "yes" if you want me to proceed with this modification.

\textbf{user}:Yes, please go ahead and make the change to the blue wireless earbuds (Item ID: 6077640618) for my order \#W5061109. Thank you for confirming the price difference and refund!

\textbf{assistant}:\{'name': 'modify\_pending\_order\_items', 'arguments': \{'order\_id': '\#W5061109', 'item\_ids': ['3694871183'], 'new\_item\_ids': ['6077640618'], 'payment\_method\_id': 'paypal\_3742148'\}\}

\textbf{tool}:\{"order\_id": "\#W5061109", "user\_id": "chen\_johnson\_4204", "address": \{"address1": "503 Elm Avenue", "address2": "Suite 641", "city": "Houston", "country": "USA", "state": "TX", "zip": "77004"\}, "items": [\{"name": "Makeup Kit", "product\_id": "5149340237", "item\_id": "6254646215", "price": 248.85, "options": \{"skin tone": "dark", "kit size": "basic", "brand": "Brand B"\}\}, \{"name": "Wireless Earbuds", "product\_id": "9924732112", "item\_id": "6077640618", "price": 242.92, "options": \{"color": "blue", "battery life": "8 hours", "water resistance": "not resistant"\}\}, \{"name": "Office Chair", "product\_id": "4794339885", "item\_id": "8323284863", "price": 511.24, "options": \{"material": "fabric", "color": "blue", "armrest": "adjustable", "backrest height": "standard"\}\}, \{"name": "Bluetooth Speaker", "product\_id": "4768869376", "item\_id": "3254583681", "price": 302.67, "options": \{"color": "blue", "battery life": "20 hours", "water resistance": "yes"\}\}], "status": "pending (item modified)", "fulfillments": [], "payment\_history": [\{"transaction\_type": "payment", "amount": 1319.43, "payment\_method\_id": "paypal\_3742148"\}, \{"transaction\_type": "refund", "amount": 13.750000000000028, "payment\_method\_id": "paypal\_3742148"\}], "cancel\_reason": null, "exchange\_items": null, "exchange\_new\_items": null, "exchange\_payment\_method\_id": null, "exchange\_price\_difference": null, "return\_items": null, "return\_payment\_method\_id": null\}

\textbf{assistant}:I've successfully modified your order. Here's the confirmation:

- Your order (\#W5061109) has been updated to include the blue wireless earbuds (Item ID: 6077640618)
- The price has been adjusted from \$256.67 to \$242.92
- A refund of \$13.75 has been processed to your PayPal account
- Your order status is now "pending (item modified)"

The rest of your order items remain unchanged:
- Makeup Kit: \$248.85
- Office Chair: \$511.24
- Bluetooth Speaker: \$302.67

Is there anything else you'd like to modify or any other assistance you need with your order?

\textbf{user}:No, that covers everything I needed. Thanks for your help!\#\#\#STOP\#\#\#

\end{tcolorbox}

\section{Details of the Evaluation Benchmarks}
\label{app:c}

\textbf{TAU1-Bench.} TAU1-Bench~\cite{yao2024tau} is a high-quality benchmark designed to evaluate the capabilities of language agents in realistic, multi-turn, tool-augmented user interaction scenarios. It focuses on assessing whether agents can follow complex domain policies, interact with real databases via tools, and maintain consistent, policy-compliant behavior throughout a dialogue. The benchmark covers two domains: retail (e-commerce customer service) and airline (airline booking and support), each equipped with realistic, modular databases and tools. Every domain's policy specifies operational constraints such as order cancellation limits, exchange/return confirmation requirements, and rules for payment methods or baggage allowances. TAU1-Bench features a collection of carefully curated 115 retail tasks and 50 airline tasks, where each user instruction is crafted to ensure a single correct outcome under the policy constraints. The tasks are diverse, demanding multi-turn interaction, tool invocation, and user confirmation. During evaluations, we used the official evaluation code.

\textbf{TAU2-Bench.} TAU2-Bench~\cite{barres2025tau} is a new benchmark designed to evaluate language agents in more realistic dual-control scenarios, building upon the foundations of TAU1-Bench. While TAU1-Bench focuses on single-agent tool use where only the agent can interact with the database, TAU2-Bench introduces TAU2 Telecom where both the user and the agent can independently perform tool calling, reflecting real-world situations such as technical support and collaborative troubleshooting. The benchmark expands the datasets by including 114 telecom tasks, in addition to retail and airline. During evaluations, we used the official evaluation code.

\textbf{Berkeley Function-Calling Leaderboard (BFCL)-V3 Multi Turn.} In addition to TAU-Bench, we also used BFCL-V3 Multi Turn~\cite{patilberkeley} 
which provides a diverse perspective for evaluating the model's ability to perform tool use.
Each category--Multi Turn Base, Miss Param, Miss Func, and Long Context--contains 200 tasks, resulting in a total of 800 tasks across all categories. The Multi Turn Base focuses on standard and diverse multi-turn interactions, providing all necessary information for task completion and requiring the model to handle user requests unambiguously. The three augmented test sets are designed to further challenge the model’s capabilities in scenarios involving missing key information, insufficient available functions, and lengthy, information-dense contexts. In our experiments, we used the official repository released by the authors and followed the provided instructions to obtain the model results.
 
\textbf{ACEBench Agent.} The Agent subset of ACEBench~\cite{chen2025acebench} is designed to evaluate the multi-turn, multi-step tool-using capabilities of LLMs in realistic user-agent collaboration scenarios. The scenarios are constructed based on real-world domains such as flight booking, food delivery, finance, and communications, and are modularized into sub-tasks reflecting authentic functional goals like user authentication, payment processing, and order management. It includes 50 tasks and 22 tools. Since both the current mainstream models and our models use the function calling (FC) mode for tool use, whereas ACEBench uses prompt evaluation by default, we modified the official evaluation code to support evaluation under the FC mode.

\end{appendices}

\end{document}